\newcommand\BestAccuracyPct{91,1\%}
\newcommand\BestAccuracyValidPct{95,6\%}

\newcommand\TrainCountAll{3401}
\newcommand\TrainCountDirected{2244}
\newcommand\TrainCountRound{1742}
\newcommand\TrainCountInstancesAll{3986}
\newcommand\ValidCountAll{528}
\newcommand\ValidCountDirected{348}
\newcommand\ValidCountRound{269}
\newcommand\ValidCountInstancesAll{617}
\newcommand\TrainValidSplit{15,5\%}
\newcommand\TestCountAll{186}
\newcommand\TestCountDirected{126}
\newcommand\TestCountRound{105}
\newcommand\TestCountInstancesAll{231}

\newcommand\DSZeroGoogle{1906}
\newcommand\DSZeroFlickr{935}
\newcommand\DSZeroFiveH{482}
\newcommand\DSZeroUn{78}
\newcommand\DSZeroDir{2244}
\newcommand\DSZeroRou{1742}
\newcommand\DSZeroSmall{685}
\newcommand\DSZeroMed{2247}
\newcommand\DSZeroLar{1054}

\newcommand\CmLiteIter{20000} 
\newcommand\CmFiveSevenIter{20000} 
\newcommand\CmNineNineIter{30000} 
\newcommand\ATSSfiftyIter{30000}
\newcommand\ATSSxIter{22500} 
\newcommand\TridentIter{20000} 

\newcommand\CmLiteTrainTime{0.28}
\newcommand\CmFiveSevenTrainTime{0.81}
\newcommand\CmNineNineTrainTime{1.22}
\newcommand\ATSSfiftyTrainTime{0.48}
\newcommand\ATSSxTrainTime{2.40}
\newcommand\TridentTrainTime{1.59}

\newcommand\CmLiteInf{0.061}
\newcommand\CmFiveSevenInf{0.089}
\newcommand\CmNineNineInf{0.109}
\newcommand\ATSSfiftyInf{0.064}
\newcommand\ATSSxInf{0.132}
\newcommand\TridentInf{0.169}

\newcommand\CmLiteAP{88,2\%}
\newcommand\CmLiteAPxx{63,8\%}
\newcommand\CmLiteAPs{65,4\%}
\newcommand\CmLiteAPm{68,7\%}
\newcommand\CmLiteARone{61,0\%}
\newcommand\CmLiteARten{73,4\%}
\newcommand\CmLiteARs{69,7\%}
\newcommand\CmLiteARm{78,8\%}

\newcommand\CmFiveSevenAP{89,7\%}
\newcommand\CmFiveSevenAPxx{62,9\%}
\newcommand\CmFiveSevenAPs{62,8\%}
\newcommand\CmFiveSevenAPm{66,3\%}
\newcommand\CmFiveSevenARone{61,5\%}
\newcommand\CmFiveSevenARten{71,5\%}
\newcommand\CmFiveSevenARs{65,7\%}
\newcommand\CmFiveSevenARm{79,5\%}

\newcommand\CmNineNineAP{90,1\%}
\newcommand\CmNineNineAPxx{64,9\%}
\newcommand\CmNineNineAPs{64,3\%}
\newcommand\CmNineNineAPm{70,2\%}
\newcommand\CmNineNineARone{61,5\%}
\newcommand\CmNineNineARten{72,9\%}
\newcommand\CmNineNineARs{68,1\%}
\newcommand\CmNineNineARm{79,7\%}

\newcommand\ATSSfiftyAP{86,5\%}
\newcommand\ATSSfiftyAPxx{58,2\%}
\newcommand\ATSSfiftyAPs{57,4\%}
\newcommand\ATSSfiftyAPm{63,9\%}
\newcommand\ATSSfiftyARone{56,8\%}
\newcommand\ATSSfiftyARten{66,0\%}
\newcommand\ATSSfiftyARs{60,7\%}
\newcommand\ATSSfiftyARm{73,5\%}

\newcommand\ATSSxAP{91,1\%}
\newcommand\ATSSxAPxx{65,4\%}
\newcommand\ATSSxAPs{64,4\%}
\newcommand\ATSSxAPm{70,1\%}
\newcommand\ATSSxARone{61,6\%}
\newcommand\ATSSxARten{72,3\%}
\newcommand\ATSSxARs{67,2\%}
\newcommand\ATSSxARm{79,4\%}

\newcommand\TridentAP{88,3\%}
\newcommand\TridentAPxx{58,7\%}
\newcommand\TridentAPs{56,0\%}
\newcommand\TridentAPm{65,4\%}
\newcommand\TridentARone{56,3\%}
\newcommand\TridentARten{65,8\%}
\newcommand\TridentARs{58,5\%}
\newcommand\TridentARm{76,0\%}

\newcommand\CmLiteVAP{93,9\%}
\newcommand\CmLiteVAPxx{69,9\%}
\newcommand\CmLiteVAPs{49,9\%}
\newcommand\CmLiteVAPm{72,1\%}
\newcommand\CmLiteVAPl{77,1\%}
\newcommand\CmLiteVARone{71,1\%}
\newcommand\CmLiteVARten{76,5\%}
\newcommand\CmLiteVARs{54,5\%}
\newcommand\CmLiteVARm{77,5\%}
\newcommand\CmLiteVARl{85,6\%}

\newcommand\CmFiveSevenVAP{95,6\%}
\newcommand\CmFiveSevenVAPxx{74,5\%}
\newcommand\CmFiveSevenVAPs{57,4\%}
\newcommand\CmFiveSevenVAPm{75,9\%}
\newcommand\CmFiveSevenVAPl{81,1\%}
\newcommand\CmFiveSevenVARone{74,6\%}
\newcommand\CmFiveSevenVARten{80,3\%}
\newcommand\CmFiveSevenVARs{60,5\%}
\newcommand\CmFiveSevenVARm{81,0\%}
\newcommand\CmFiveSevenVARl{88,0\%}

\newcommand\CmNineNineVAP{95,0\%}
\newcommand\CmNineNineVAPxx{73,3\%}
\newcommand\CmNineNineVAPs{55,5\%}
\newcommand\CmNineNineVAPm{74,0\%}
\newcommand\CmNineNineVAPl{82,1\%}
\newcommand\CmNineNineVARone{74,0\%}
\newcommand\CmNineNineVARten{79,5\%}
\newcommand\CmNineNineVARs{58,8\%}
\newcommand\CmNineNineVARm{79,4\%}
\newcommand\CmNineNineVARl{89,3\%}

\newcommand\ATSSfiftyVAP{93,2\%}
\newcommand\ATSSfiftyVAPxx{71,7\%}
\newcommand\ATSSfiftyVAPs{52,3\%}
\newcommand\ATSSfiftyVAPm{74,4\%}
\newcommand\ATSSfiftyVAPl{77,6\%}
\newcommand\ATSSfiftyVARone{72,9\%}
\newcommand\ATSSfiftyVARten{78,2\%}
\newcommand\ATSSfiftyVARs{55,4\%}
\newcommand\ATSSfiftyVARm{80,0\%}
\newcommand\ATSSfiftyVARl{86,3\%}

\newcommand\ATSSxVAP{94,3\%}
\newcommand\ATSSxVAPxx{73,1\%}
\newcommand\ATSSxVAPs{54,6\%}
\newcommand\ATSSxVAPm{76,0\%}
\newcommand\ATSSxVAPl{78,0\%}
\newcommand\ATSSxVARone{74,2\%}
\newcommand\ATSSxVARten{79,6\%}
\newcommand\ATSSxVARs{57,8\%}
\newcommand\ATSSxVARm{81,2\%}
\newcommand\ATSSxVARl{86,8\%}

\newcommand\TridentVAP{94,0\%}
\newcommand\TridentVAPxx{70,3\%}
\newcommand\TridentVAPs{52,7\%}
\newcommand\TridentVAPm{70,9\%}
\newcommand\TridentVAPl{79,8\%}
\newcommand\TridentVARone{73,0\%}
\newcommand\TridentVARten{77,6\%}
\newcommand\TridentVARs{55,6\%}
\newcommand\TridentVARm{78,7\%}
\newcommand\TridentVARl{86,8\%}

\newcommand\CmLiteClass{0.007}
\newcommand\CmLiteLoc{0.048}
\newcommand\CmLiteCenter{0.595}
\newcommand\CmLiteMask{0.079}
\newcommand\CmLiteTotal{0.732}

\newcommand\CmFiveSevenClass{0.009}
\newcommand\CmFiveSevenLoc{0.041}
\newcommand\CmFiveSevenCenter{0.593}
\newcommand\CmFiveSevenMask{0.071}
\newcommand\CmFiveSevenTotal{0.715}

\newcommand\CmNineNineClass{0.007}
\newcommand\CmNineNineLoc{0.031}
\newcommand\CmNineNineCenter{0.591}
\newcommand\CmNineNineMask{0.057}
\newcommand\CmNineNineTotal{0.692}

\newcommand\ATSSfiftyClass{0.003}
\newcommand\ATSSfiftyLoc{0.045}
\newcommand\ATSSfiftyCenter{0.594}
\newcommand\ATSSfiftyTotal{0.645}

\newcommand\ATSSxClass{0.015}
\newcommand\ATSSxLoc{0.052}
\newcommand\ATSSxCenter{0.593}
\newcommand\ATSSxTotal{0.663}

\newcommand\TridentClass{0.010}
\newcommand\TridentLoc{0.007}
\newcommand\TridentTotal{0.052}

\newcommand\CmLiteFile{281.1}
\newcommand\CmFiveSevenFile{551.4}
\newcommand\CmNineNineFile{776.0}
\newcommand\ATSSfiftyFile{264.1}
\newcommand\ATSSxFile{718.7}
\newcommand\TridentFile{421.9}

\newcommand\DSOneTrainCountAll{2457}
\newcommand\DSOneTrainCountDirected{1554}
\newcommand\DSOneTrainCountRound{1547}
\newcommand\DSOneTrainInstances{3101}
\newcommand\DSOneValidCountAll{270}
\newcommand\DSOneValidCountDirected{178}
\newcommand\DSOneValidCountRound{176}
\newcommand\DSOneValidInstances{354}

\newcommand\DSOneGoogle{2457}
\newcommand\DSOneDir{1554}
\newcommand\DSOneRou{1547}
\newcommand\DSOneSmall{762}
\newcommand\DSOneMed{2147}
\newcommand\DSOneLar{193}

\newcommand\DSTwoCountAll{4167}
\newcommand\DSTwoInstances{5380}
\newcommand\DSTwoCountDirected{3325}
\newcommand\DSTwoCountRound{2055}

\newcommand\DSAllTrainCountAll{8387}
\newcommand\DSAllTrainCountDirected{6137}
\newcommand\DSAllTrainCountRound{4282}
\newcommand\DSAllTrainCountInstancesAll{10419}
\newcommand\DSAllValidCountAll{533}
\newcommand\DSAllValidCountDirected{379}
\newcommand\DSAllValidCountRound{268}
\newcommand\DSAllValidCountInstancesAll{647}

\newcommand\DSAllGoogle{6598}
\newcommand\DSAllBaidu{269}
\newcommand\DSAllFlickr{960}
\newcommand\DSAllFiveH{482}
\newcommand\DSAllUn{78}

\newcommand\DSAllSmall{2331}
\newcommand\DSAllMed{6397}
\newcommand\DSAllLar{1691}

\newcommand\ResTestEightAP{92,0\%}
\newcommand\ResTestEightAPbig{71,4\%}
\newcommand\ResTestEightAPm{91,5\%}
\newcommand\ResTestEightAPl{96,0\%}
\newcommand\ResTestEightAR{94,5\%}
\newcommand\ResTestEightARm{93,7\%}
\newcommand\ResTestEightARl{100\%}
\newcommand\ResTestEightF{93,2\%}
\newcommand\ResTestEightDir{89,2\%}
\newcommand\ResTestEightRou{94,8\%}

\newcommand\ResValEightAP{97,3\%}
\newcommand\ResValEightAPbig{80,1\%}
\newcommand\ResValEightAPm{96,8\%}
\newcommand\ResValEightAPl{98,7\%}
\newcommand\ResValEightAR{98,4\%}
\newcommand\ResValEightARm{98,0\%}
\newcommand\ResValEightARl{99,6\%}
\newcommand\ResValEightF{97,8\%}
\newcommand\ResValEightDir{98,0\%}
\newcommand\ResValEightRou{96,5\%}

\newcommand\DetTestEightAP{91,5\%}
\newcommand\DetTestEightAPbig{68,9\%}
\newcommand\DetTestEightAPm{91,1\%}
\newcommand\DetTestEightAPl{93,9\%}
\newcommand\DetTestEightAR{93,1\%}
\newcommand\DetTestEightARm{92,6\%}
\newcommand\DetTestEightARl{95,5\%}
\newcommand\DetTestEightF{92,3\%}
\newcommand\DetTestEightDir{89,1\%}
\newcommand\DetTestEightRou{93,9\%}

\newcommand\DetValEightAP{98,7\%}
\newcommand\DetValEightAPbig{83,4\%}
\newcommand\DetValEightAPm{99,2\%}
\newcommand\DetValEightAPl{98,8\%}
\newcommand\DetValEightAR{99,5\%}
\newcommand\DetValEightARm{99,7\%}
\newcommand\DetValEightARl{99,1\%}
\newcommand\DetValEightF{98,5\%}
\newcommand\DetValEightDir{98,6\%}
\newcommand\DetValEightRou{98,7\%}

\newcommand\ResTestEightAPS{91,5\%}
\newcommand\ResTestEightAPbigS{70,2\%}
\newcommand\ResTestEightAPmS{90,9\%}
\newcommand\ResTestEightAPlS{96,2\%}
\newcommand\ResTestEightARS{94,0\%}
\newcommand\ResTestEightARmS{93,2\%}
\newcommand\ResTestEightARlS{100\%}
\newcommand\ResTestEightFS{92,7\%}
\newcommand\ResTestEightDirS{89,3\%}
\newcommand\ResTestEightRouS{93,8\%}

\newcommand\ResValEightAPS{97,0\%}
\newcommand\ResValEightAPbigS{80,8\%}
\newcommand\ResValEightAPmS{96,5\%}
\newcommand\ResValEightAPlS{98,8\%}
\newcommand\ResValEightARS{98,1\%}
\newcommand\ResValEightARmS{97,5\%}
\newcommand\ResValEightARlS{99,6\%}
\newcommand\ResValEightFS{97,5\%}
\newcommand\ResValEightDirS{97,3\%}
\newcommand\ResValEightRouS{96,8\%}

\newcommand\ResTestTwelveAP{90,1\%}
\newcommand\ResTestTwelveAPbig{71,9\%}
\newcommand\ResTestTwelveAPm{90,8\%}
\newcommand\ResTestTwelveAPl{87,0\%}
\newcommand\ResTestTwelveAR{94,0\%}
\newcommand\ResTestTwelveARm{93,7\%}
\newcommand\ResTestTwelveARl{95,5\%}
\newcommand\ResTestTwelveF{92,0\%}
\newcommand\ResTestTwelveDir{88,3\%}
\newcommand\ResTestTwelveRou{92,0\%}

\newcommand\DetTestTwelveAP{92,6\%}
\newcommand\DetTestTwelveAPbig{72,8\%}
\newcommand\DetTestTwelveAPm{92,4\%}
\newcommand\DetTestTwelveAPl{93,3\%}
\newcommand\DetTestTwelveAR{95,4\%}
\newcommand\DetTestTwelveARm{95,3\%}
\newcommand\DetTestTwelveARl{95,5\%}
\newcommand\DetTestTwelveF{94,0\%}
\newcommand\DetTestTwelveDir{90,0\%}
\newcommand\DetTestTwelveRou{95,2\%}

\documentclass[conference]{IEEEtran}
\IEEEoverridecommandlockouts
\usepackage{cite}
\usepackage{amsmath,amssymb,amsfonts}
\usepackage{algorithmic}
\usepackage{graphicx}
\usepackage{textcomp}
\usepackage{xcolor}

\usepackage{subfigure}

\usepackage{kantlipsum} 
\usepackage{mwe} 

\usepackage{tikz}
\usepackage{amsmath}

\usepackage{filecontents}

\usepackage[square,sort,comma,numbers]{natbib}

\usepackage[font=scriptsize,labelfont=bf]{caption}

\usepackage{url}

\usepackage{cite}
\usepackage{amsmath,amssymb,amsfonts}
\usepackage{algorithmic}
\usepackage{graphicx}
\usepackage{textcomp}
\usepackage{xcolor}

\usepackage{subfigure}

\usepackage{kantlipsum} 
\usepackage{mwe} 

\newcommand{\specialcell}[2][c]{%
	\begin{tabular}[#1]{@{}c@{}}#2\end{tabular}}

\usepackage{color}
\usepackage{framed}

\newcounter{t0d0_counter}
\newcommand{\nofixme}[1]{
}
\newcommand{\fixme}[1]{
  \stepcounter{t0d0_counter}
  \definecolor{shadecolor}{rgb}{1,1,0} 
  \begin{shaded}
  T0D0 \arabic{t0d0_counter}: #1
  \end{shaded}
}

\usepackage{graphicx}
\usepackage{comment}
\usepackage{capt-of}
\usepackage{booktabs}
\usepackage{hyperref}
\usepackage{listings}
\usepackage{balance}
\usepackage{algorithm,algorithmic}
\usepackage{enumitem}
\setlist{nolistsep}
\setlist[itemize]{leftmargin=*}

\usepackage{array}
\newcolumntype{H}{>{\setbox0=\hbox\bgroup}c<{\egroup}@{}}

\usepackage{grffile}

\def\BibTeX{{\rm B\kern-.05em{\sc i\kern-.025em b}\kern-.08em
    T\kern-.1667em\lower.7ex\hbox{E}\kern-.125emX}}
\begin{document}

\title{
Towards large-scale, automated, accurate detection of CCTV camera objects using computer vision \\ 
{\large Applications and implications for privacy, anonymity, surveillance, safety, and cybersecurity \\ (Preprint)}
}

\author{
\IEEEauthorblockN{
Hannu Turtiainen\IEEEauthorrefmark{5} \textsuperscript{$\dagger$},
Andrei Costin\IEEEauthorrefmark{5} \textsuperscript{$\circledast$}, 
Tuomo Lahtinen\IEEEauthorrefmark{4}, \\
Lauri Sintonen\IEEEauthorrefmark{4}, and
Timo H\"am\"al\"ainen\IEEEauthorrefmark{5}
}
\IEEEauthorblockA{
\textit{University of Jyv\"askyl\"a}\\
Jyv\"askyl\"a, Finland \\
\IEEEauthorrefmark{5}\{turthzu,ancostin,timoh\}@jyu.fi, \\
\IEEEauthorrefmark{4}\{tuomo.t.lahtinen,lauri.m.j.sintonen\}@student.jyu.fi
}
\thanks{$\dagger$ This paper is based on author's MSc thesis~\cite{Turtiainen:Thesis:2020}.}
\thanks{$\circledast$ Corresponding and original idea's author.}
}

\maketitle

\begin{abstract}
%
%
While the earliest known CCTV camera was developed almost a century ago back in 1927, 
currently, it is assumed as granted there are about 770 millions CCTV cameras 
around the globe, and their number is casually predicted to surpass 1 billion in 2021. 
Similarly to the first prototypes from 1927, at present the main promoted 
benefits for using and deploying CCTV cameras are physical security, 
safety, and prophylactics of crime. 
At the same time the increasing, widespread, unwarranted, and unaccountable 
use of CCTV cameras globally raises privacy risks and concerns for the 
last several decades. Recent technological advances implemented in CCTV 
cameras such as AI-based facial recognition and IoT connectivity only 
fuel further concerns raised by privacy advocates. 

In order to withstand the ever-increasing invasion of privacy by 
CCTV cameras and technologies, on par \emph{CCTV-aware solutions} must 
exist that provide privacy, safety, and cybersecurity features. 
We argue that a first important step towards such CCTV-aware solutions 
must be a mapping system (e.g., Google Maps, OpenStreetMap) that provides 
both privacy and safety routing and navigation options. 
However, this in turn requires that the mapping system contains updated 
information on CCTV cameras' exact geo-location, coverage area, and possibly 
other meta-data (e.g., resolution, facial recognition features, operator). 
Such information is however missing from current mapping systems, 
and there are several ways to fix this. 
One solution is to perform CCTV camera detection on geo-location tagged 
images, e.g., street view imagery on various platforms, user images publicly 
posted in image sharing platforms such as Flickr. 
Additionally, mobile devices, equipped with video input and GPS, 
can be designed and built that can aid real-time data collection and 
mapping of CCTV cameras in the real-world. 
At present, the only fast, scalable and feasible way to achieve these 
is to apply computer vision object detection techniques. 
Unfortunately, to the best of our knowledge, there are no computer vision 
models for CCTV camera object detection as well as no mapping system 
that supports privacy and safety routing options. 

To close these gaps, with this paper we introduce \emph{CCTVCV} -- the first and only 
computer vision MS COCO-compatible models that are able to accurately 
detect CCTV and video surveillance cameras in images and video frames. 
%
To this end, our best detectors were built using \DSAllTrainCountAll{} 
images that were manually reviewed and annotated to contain 
\DSAllTrainCountInstancesAll{} CCTV camera instances, 
and achieve an accuracy of up to \DetValEightAP{}. 
Moreover, we build and evaluate multiple models, present a comprehensive 
comparison of their performance, and outline core challenges associated 
with such research. 
We also present possible privacy-, safety-, and security-related practical 
applications of our core work, including \emph{previews and excerpts} from our 
working prototypes of CCTV-aware privacy and safety routing and navigation. 
Last but not least, we release as open-data and open-source relevant 
data and code that can be used to validate and further extend our work. 


\end{abstract}

\section{Introduction}
\label{sec:intro}

CCTV and video surveillance cameras represent nowadays some of the 
most ubiquitous technology, and it is almost impossible to live a day 
without getting into the field of view of at least one, if not dozens of, CCTV camera(s)~\cite{bi2019cctv,barrett2013one}. 
At present, CCTV cameras are an integral part of any infrastructure 
(e.g., cities, buildings, streets, businesses), and it is expected that 
by 2021 there will be more than 1 billion CCTV cameras 
globally~\cite{cnbc2019billion}. 
Meanwhile, the earliest known precursor of a modern CCTV camera goes back as early as 1927 when 
the Soviet inventor Leon Theremin installed the first real-world usable 
prototypes of then-called \emph{distance vision} along the 
Kremlin premises~\cite{glinsky2000theremin}. It was a mechanically-operated 
device that transmitted a few hundred image-lines which allowed its operators to distinguish and even recognize faces. 

In terms of cybersecurity, CCTV cameras, DVRs, and video surveillance systems are already known to be the subject of numerous 
cyberattacks~\cite{costin2016security}, and they were also the main 
culprit behind the now legendary and massive attack 
by the Mirai IoT botnet~\cite{antonakakis2017understanding}. 
At the same time, it is long and well known that CCTV cameras raise concerns 
and pose risks related to privacy~\cite{eff-cctv,slobogin2002public,larsen2011setting,ryberg2007privacy,goold2002privacy}.
However, it is very hard (if not impossible) at present to accurately and 
objectively assess and address the privacy risks and implications. 

There are several ways to mitigate the privacy risks posed by CCTV cameras 
(including their additional features such as face recognition).
One possible and commonly used and promoted method is to use artistic (but perhaps unpractical and \emph{low-tech}) 
approaches such as specially-designed transparent plastic masks~\cite{ft20dazzle} 
or face painting (i.e., ``adversarial computer vision'' attack)~\cite{vice20makeup}. 
However, these methods could be easily defeated as the advances in computer 
vision and face recognition are extremely fast, allowing correct 
identification through face recognition even when the subjects wear 
respiratory masks~\cite{reut20mask}. 

Another possible way is to develop, provide and use appropriate 
\emph{high-tech tools} against the invasion of privacy by CCTV cameras. 
Examples of such tools include CCTV-aware route planning and navigation, 
and real-time early warning system when mobile and embedded 
devices that are video-input equipped (e.g., wearables, smartphones, drones) 
enter areas under the potential field of view of CCTV cameras.
To this end, such tools require trustworthy object detection and counting, 
and accurate mapping and localization. 
In this context, computer vision is a proven method that excellently 
performs for object detection and counting~\cite{onoro2016towards}, 
as well as for mapping and localization~\cite{fuentes2015visual,verhoeven2012mapping}.
However, in order to implement such CCTV-aware technology, 
various foundational blocks are currently missing. One such foundation block 
of utmost importance are 
object detectors for quick, accurate, and automated computer vision detection 
of CCTV cameras.

\subsection{Extra motivation}
\label{sec:anecdot}

Finally, in order to assess how serious is a privacy risk of a certain 
local/global CCTV installation is, certain modeling approaches require 
to identify and precisely count the CCTV cameras. 
Those models also require to know exactly where the cameras are located along with their 
other characteristics such as field of view, zoom levels, and other 
built-in features (e.g., face recognition, Infra-Red (IR), Pan-Tilt-Zoom (PTZ)). 
Unfortunately, most of the current data publicly available about 
CCTV cameras statistics and characteristics, both at global and local levels, 
can be considered \emph{completely unreliable}. 
As follows, we provide several such examples 
that demonstrate the unsound methodologies and discrepancies in data. 
In one instance, the UK had until 2014 three different major estimations about 
the number of CCTV cameras -- 
1.8 mil.~\cite{gerrard2011two}, 
4.2 mil.~\cite{mccahill2002cctv}, 
and 5.9 mil.~\cite{barrett2013one}. 
To add insult to the injury, 
despite the rampant increase of CCTV and video surveillance globally, 
these numbers were not updated ever since, and are referenced in 
2019--2020 as ``current'' by various reports and media outlets. 
In another instance, the estimates for global number of CCTV cameras vary 
between 25 mil.~\cite{caughtoncamera2019cctvlond} and 
770 mil.~\cite{cnbc2019billion} -- a whopping 25x discrepancy. 
In yet another instance, the UK finds that on average a person enters 
a CCTV camera view 300 times a day~\cite{caughtoncamera2019cctvlond}. 
A similar study in the US puts that number at 50+ times a day, despite the 
more worrying fact that the average US respondent \emph{assumed 
it was 4 cameras or less}~\cite{ipvm2016us} -- at least a 10x lower 
presumed privacy risk and exposure than in reality. 
At the same time, a recent journalist experiment in NYC (US) by 
Pasley~\cite{bi2019cctv} found he encountered CCTV cameras face-to-face 
at least 49 times, and that is 
\emph{just counting a single trip to the workplace}.
Finally, there are also discrepancies related to the number of cameras per 
1000 persons~\cite{2019citymostcctv,barrett2013one,gerrard2011two}. 
%
%
A quick check reveals that such discrepancies may have several root-causes. 
In some cases, it is the use of unsound and low-tech methods, such as visually 
counting CCTV cameras on a \emph{single main shopping street in London}, 
and then extrapolating (by some unvetted model) the numbers to the entire 
country~\cite{mccahill2002cctv}. 
In other cases, it is the heavy use of sales and marketing 
data~\cite{cnbc2019billion}, which by our experience very often is 
unrepresentative, highly approximated, and over-estimated. 
Even if we would assume the rightfulness of counting data provided by such 
unscientific methods, they cannot however provide the privacy-critical 
information about any camera, namely its location, characteristics 
(e.g., the field of view, zoom), and spatial coverage. 

\subsection{Contributions}
\label{sec:contrib}

In this paper, we try to close the existing fundamental research and 
technology gaps 
as well as to address the strong and imminent need for such tools. 
%
%
During the experiments on real-world data, our system achieved an 
accuracy of up to \DetValEightAP{}, 
which is comparable to Google's original automatic system for large-scale 
privacy protection of human faces and car license plates in 
Google Street View~\cite{frome2009large}.

\begin{itemize}
\itemsep0em

\item We are the first to research, implement and evaluate 
Computer Vision (CV) models -- \emph{CCTVCV} -- to detect 
\emph{CCTV camera objects}~\footnote{Our core aim at this stage is to be able 
to accurately detect \emph{generally visible} CCTV cameras at a large-scale. 
Although our object detectors work well on certain edge-cases 
(Figures~\ref{fig:img_4},~\ref{fig:img_9}), the camouflage attacks 
on object detectors (e.g., careful placement, decoration) is a separate emerging topic~\cite{huang2020universal}.}
in images and video frames, with \emph{particular focus on privacy and anonymity applications}.

\item We are first to introduce and motivate a handful of \emph{CCTV-aware} 
applications and \emph{working prototypes} for both \emph{mobile devices} and 
\emph{privacy-first/safety-first routing} scenarios as relevant for modern digitized lifestyle. 

\item We release \emph{CCTVCV} as \emph{open data} the models and datasets 
necessary to validate our results and to further expand the datasets 
and the research field. 
To our knowledge, these are the first, the largest, and the best-performing 
datasets and models to be publicly released for solving the stated problems. 

\item The relevant artefacts (e.g., code, datasets, trained models, and documentation) will be available at: 
\textit{\url{https://github.com/Fuziih}} and 
\textit{\url{https://etsin.fairdata.fi/dataset/d2d2d6e2-0b5c-46e0-8833-53d8a24838a0} (urn:nbn:fi:att:258ce5ad-9501-46b9-a707-c1f59689ee10)}. 

\end{itemize}

\subsection{Paper organization}
\label{sec:org}

The rest of this paper is organized as follows.
We detail our methodology and describe our experimental setup in Section~\ref{sec:method}. 
Then, in Section~\ref{sec:results} we present our results and main findings. 
In Section~\ref{sec:discuss} we discuss the caveats, possible applications, and future work. 
In Section~\ref{sec:relwork} we overview the related work.
Finally, we conclude with Section~\ref{sec:concl}. 

\section{Methodology and experimental setup}
\label{sec:method}


As with any Computer Vision (CV) object detector, we followed a two-phase 
approach. First, we trained multiple models for object detection using 
``training sets'' (and additional ``validation sets'' for internal 
self-validation during model's training). 
In order to train the CV object detectors, we split the phase into four 
parts -- 
dataset gathering (Section~\ref{sec:dataset}), 
image annotation (Section~\ref{sec:imageannot}), 
environment setup (Section~\ref{sec:envsetup}), 
and model training (Section~\ref{sec:modeltrain}). 
Once the model training is completed, we evaluate each trained model against a ``testing set''. 
Each dataset (see below, e.g., \texttt{Dataset0}, \texttt{DatasetAll}) 
had its own ``testing set'' which was withheld from the training and 
used only on finalized models as ``in the wild'' testing. 

\subsection{Dataset gathering and definitions}
\label{sec:dataset}

When we started to gather the dataset, we made a practical decision 
that we want to be able to classify the cameras into at least two distinct 
sub-classes based on their shape -- \emph{directed cameras} and \emph{round cameras}. 
Having this information allows us in the future work to model more accurately 
their field of view coverage in 3D, therefore allowing to decide whether 
a particular point in space (e.g., sidewalk, street, street corner) provides 
or not privacy to a person. 
However, it is important to note that, from the point of view of MS COCO 
annotations~\cite{lin2014microsoft} and CV/ML training, we treat all the 
CCTV cameras as a single \emph{camera} category, regardless of whether they 
are annotated as directed or round. 
This is in line with the MS COCO~\cite{lin2014microsoft} approach, where 
for example, the \emph{cars} category is represented as a single category 
regardless of the actual cars' properties (e.g, shape, make, color).


\emph{Directed cameras} include box- and bullet-shaped cameras 
(see Figure~\ref{fig:sample_directed}), and we assume such cameras record 
a limited field of view specifically in the direction they are pointed to. 
Some of them may be motorized (e.g., via Pan-Tilt-Zoom (PTZ) hardware 
and protocols) and therefore be able to have a mobile 
field of view (in theory up to 360$^{\circ}$). 
However, it is challenging (if not impossible) to detect PTZ with 
computer vision on static (and low resolution) images. 
Therefore, to simplify a bit our experiments and future geo-mapping modeling, 
we assume that such cameras are static, cover the particular direction 
they are pointed to and have limited vertical and horizontal fields of view. 

\emph{Round cameras} include dome- and sphere-shaped cameras 
(see Figure~\ref{fig:sample_round}), and we assume such cameras potentially 
record a 360$^{\circ}$ field of view. 
Even though some dome- and sphere-shaped cameras host inside a static and 
directed camera sensor, most of the times it is challenging to know that 
because of the reflective glass. 
Therefore, to simplify a bit our experiments and future geo-mapping modeling, 
we assume that such cameras record a 360$^{\circ}$ field of view. 

To date, the collection, annotation, and quality-check of all the 
datasets presented in this paper took the equivalent of 
at least \emph{five and a half person-months of effort}.

\subsection{Datasets overview}
\label{sec:datasetoverview}

\paragraph{Dataset0.}
Our first dataset, that we reference as \texttt{Dataset0}, was originally collected 
during January 2020 and February 2020 by contributor \emph{Person0 HT} and later appended 
with more images from contributor \emph{Person6 TL}. The images were collected from 
Flickr~\cite{flickr}, 500px~\cite{500px}, Unsplash~\cite{unsplash} and 
Google Maps Street View~\cite{googlesv}. 
All the images in this dataset were annotated using Wada's Labelme 
standalone annotation tool~\cite{wada_labelme} by contributor \emph{Person0 HT}. 
The \texttt{Dataset0} training set contained $\TrainCountAll$ images 
with $\TrainCountInstancesAll$ instances ($\TrainCountDirected$ directed 
and $\TrainCountRound$ round) and the validation set had $\ValidCountAll$ 
images with $\ValidCountInstancesAll$ instances ($\ValidCountDirected$ 
directed and $\ValidCountRound$ round) respectively. 
The \texttt{Dataset0} also featured a small testing set of $\TestCountAll$ 
images with $\TestCountInstancesAll$ instances ($\TestCountDirected$ directed 
and $\TestCountRound$ round)~\footnote{Later on, this testing set was however replaced with improved alternatives.}.

For the \texttt{Dataset0} we trained the following models: 
Centermask2~\cite{lee_park_cm} (backbone variants: VoVNet V2 Lite-39, 50 and 99), 
ATSS~\cite{zhang_atss} (backbone variants: ResNet-50, ResNeXt-101), 
and TridentNet~\cite{li_trident} (backbone variant: ResNet-101).
Excellent performance of this early and original \texttt{Dataset0} 
could already be observed, as detailed in Appendix in 
Tables~\ref{tab:ds0-tab2},~\ref{tab:ds0-tab3},~\ref{tab:ds0-tab1}.

After several rounds of peer-review rejections and comments, it became clear 
that the security and privacy community insists our \texttt{Dataset0}, is too 
small (hence not representative) despite its novelty and good performance. 
Therefore, we set ourselves on a quest to collect a dataset large enough 
that reaches, or even exceeds, the median size for a given object category 
as set by state-of-the-art computer vision works. 
MS COCO is a definitive state-of-the-art work which is a reference point 
for datasets related to object detection by computer vision. 
It contains 80 annotated object categories representing 860001 annotated 
object instances. In MS COCO, the median size of the training set 
per detected object category is 6097. 
Therefore, as detailed below we collected several more incremental datasets 
atop \texttt{Dataset0}.

\paragraph{Dataset1.}
As our most immediate goal is to detect cameras from street-level imagery, 
our next effort to improve the dataset was to focus on dataset's content 
to street-level imagery. 
Therefore, \texttt{Dataset1} only included the street view images from \texttt{Dataset0} 
with some new images from contributor \emph{Person6 TL} that 
were captured during May 2020 from Google Street View~\cite{googlesv}. 
All the new additions were also annotated using Wada's Labelme 
standalone annotation tool~\cite{wada_labelme} by contributor \emph{Person0 HT}. 
The \texttt{Dataset1} training set contained $\DSOneTrainCountAll$ images 
with $\DSOneTrainInstances$ instances ($\DSOneTrainCountDirected$ directed 
and $\DSOneTrainCountRound$ round) and the validation set 
had $\DSOneValidCountAll$ images containing $\DSOneValidInstances$ 
instances ($\DSOneValidCountDirected$ directed and $\DSOneValidCountRound$ round). 
The testing set stayed the same as with \texttt{Dataset0}.

\paragraph{Dataset2.}
The \texttt{Dataset2} was collected during late September 2020 using 
a crowd-sourcing effort by eight (8) contributors. For this, we used a custom 
developed annotation tool (see Section~\ref{sec:imageannot}).
This dataset contains $\DSTwoCountAll$ images with $\DSTwoInstances$ camera 
instances ($\DSTwoCountDirected$ directed and $\DSTwoCountRound$ round). 
The images were captured mostly from Google Street View~\cite{googlesv}, 
but Flickr~\cite{flickr} and Baidu Maps~\cite{baidu-total-view} were also used.
This dataset was never used \emph{``as is''} for model training and evaluation. 
However, all of the images captured in this dataset were used to create the training, 
validation and testing sets of our most complete dataset \texttt{DatasetAll} that we detail below.

\paragraph{DatasetAll.}
Our latest and most complete dataset is \texttt{DatasetAll}. 
It was obtained by ``merging'' \texttt{Dataset0}, \texttt{Dataset1}, \texttt{Dataset2}, 
and applying de-duplication, cleanup and quality check after the merging process.
It is important to note that due to the complex nature of the above process, 
the counts in \texttt{DatasetAll} are not just a plain sum-up of all the 
individual counts from \texttt{Dataset0}, \texttt{Dataset1}, \texttt{Dataset2}. 
After the ``merging'' process is completed, the \texttt{DatasetAll} 
training set contains $\DSAllTrainCountAll$ images 
with $\DSAllTrainCountInstancesAll$ camera instances ($\DSAllTrainCountDirected$ directed 
and $\DSAllTrainCountRound$ round), while its validation set 
has $\DSAllValidCountAll$ images with $\DSAllValidCountInstancesAll$ camera 
instances ($\DSAllValidCountDirected$ directed and $\DSAllValidCountRound$ round). 
We used \texttt{DatasetAll} to train our most recent models, namely 
``ResNeSt-200'' and ``DetectoRS: Cascade + ResNet-50''.

\subsection{Image annotation}
\label{sec:imageannot}


Since we use supervised training at this stage, we have to label and annotate 
the images in our datasets.

\paragraph{Dataset0, Dataset1.}
To annotate the images in these datasets, we used standalone deployed Wada's 
Labelme tool~\cite{wada_labelme}. Labelme is heavily inspired by the work 
by Russell et al.~\cite{russell_labelme} who created LabelMe -- 
 web-based image annotation and labeling tool that operates in the browser 
but which requires the images to be uploaded and managed on own backend server, 
i.e., cannot annotate the images directly displayed in the browser while 
browsing third-party sites. 
Wada's Labelme~\cite{wada_labelme} allows annotation with polygon segments 
that can be used in object segmentation architectures such as Mask-RCNN \cite{he_mask} and CenterMask \cite{lee_park_cm}.
When using object segmentation, the outlines of the objects can be identified 
more precisely instead of a mere bounding box around the object of interest. 
Also, Wada's Labelme outputs individual JSON files for each annotated image, 
and includes Python scripts to embed the data as a single annotation file 
in MS COCO format~\cite{wada_labelme}.

\paragraph{Dataset2.}
Standalone tools such as Wada's Labelme~\cite{wada_labelme} come with
high-overhead and high-maintenance costs, and are not the best suited 
for distributed crowd-sourcing efforts.
To enable fast and easy crowd-source contribution, we have designed and 
developed from scratch a novel annotation tool implemented as 
an extremely flexible and COCO-compatible \emph{browser-only extension}. 
Aiming at one-click solutions, it requires minimal setup and configuration 
effort, and is ready to use out of the box even by novice contributors.
The browser extension is written in JavaScript and comes with a set of 
more than 10 distinctive features.
To date, it took the equivalent of at least \emph{five person-months of effort} 
to design, develop, test, and improve our browser-only annotation tool. 
As part of a separate publication, we describe 
the implementation details as well as its validation results, and we are 
releasing the tool under the open-source license.

\paragraph{Post-annotation.}
Where needed and applicable, we annotated the final versions of our datasets 
(i.e., training, validation, testing) with polygon shapes, and converted 
those datasets to MS COCO-compatible format. 
We also saved the individual JSON annotation files 
for future reference. For example, they may be useful when annotation changes 
are needed, or when a different splitting of the dataset into training 
and validation subsets is required.

\subsection{Datasets details}

\begin{table}[htb]
\centering
\caption{Comparison: our datasets vs. MS COCO 2017.}
\label{tab:DatasetAll-vs-MsCoco}
\resizebox{\columnwidth}{!}{%
  \begin{tabular}{lrrrrr}
    \toprule
    \specialcell{Dataset Name} & \specialcell{Total \\ Categories} & \specialcell{Total \\ Instances} & \specialcell{Median \\ Instances \\ per categ.} & \specialcell{Increase \\ (vs. Median \\ MS COCO)} \\

    \midrule
    \specialcell{MS COCO (train)}    &   80   &   860001    &   6097    & 1x \\
    \specialcell{\texttt{DatasetAll} (train)}      &   1     &   \DSAllTrainCountInstancesAll            &   \textbf{\DSAllTrainCountInstancesAll}   & \textbf{1.70x} \\
    \specialcell{\texttt{Dataset0} (train)}      &   1     &   \TrainCountInstancesAll            &   \TrainCountInstancesAll   & 0.65x \\

    \midrule
    \specialcell{MS COCO (val)}    &   80   &   36781    &   265    & 1x \\
    \specialcell{\texttt{DatasetAll} (val)}      &   1     &   \DSAllValidCountInstancesAll            &   \textbf{\DSAllValidCountInstancesAll}   & \textbf{2.44x} \\
    \specialcell{\texttt{Dataset0} (val)}      &   1     &   \ValidCountInstancesAll            &   \ValidCountInstancesAll   & 2.32x \\

	\bottomrule
  \end{tabular}
}
\end{table}

\begin{table}[htb]
\centering
\caption{High-level statistics for the datasets (training set).}
\label{tab:dataset2-images}
\resizebox{\columnwidth}{!}{%
    \begin{tabular}{l|r|r|r|r}%
    \toprule

    \multicolumn{1}{c|}{}   &   \texttt{Dataset0}    &   \texttt{Dataset1}    &   \texttt{Dataset2}    &   \texttt{DatasetAll}    \\
    \midrule
    
    \multicolumn{1}{c|}{Total counts}       &   \multicolumn{4}{r}{}    \\
    \midrule
    \specialcell{Total collected images}                           &   $\TrainCountAll$    &   $\DSOneTrainCountAll$     &   $\DSTwoCountAll$    &   $\DSAllTrainCountAll$    \\
    \specialcell{Total annotated camera instances}                 &   $\TrainCountInstancesAll$    &   $\DSOneTrainInstances$    &   $\DSTwoInstances$    &   $\DSAllTrainCountInstancesAll$    \\

    \midrule

    \multicolumn{1}{c|}{Images grouped by source}       &   \multicolumn{4}{r}{}    \\
    \midrule
    \specialcell{Google (Street View, \\ Images Search)}    &   \DSZeroGoogle    &    \DSOneGoogle   &   3873    &   \DSAllGoogle    \\
    \specialcell{Baidu street view}                     &   -    &   -    &   269    &   \DSAllBaidu    \\
    \specialcell{Flickr}                                &   \DSZeroFlickr    &   -    &   25    &   \DSAllFlickr    \\
    \specialcell{500px}                                 &   \DSZeroFiveH     &   -    &     -
        & \DSAllFiveH \\
    \specialcell{Unsplash}                              &   \DSZeroUn       &   -    &  -
        & \DSAllUn \\

    \midrule

    \multicolumn{1}{c|}{Instances grouped by sub-type}       &   \multicolumn{4}{r}{}    \\
    \midrule
    \specialcell{Directed camera instances}             &   \DSZeroDir    &   \DSOneDir    &   $\DSTwoCountDirected$    &   \DSAllTrainCountDirected    \\
    \specialcell{Round camera instances}                &   \DSZeroRou    &  \DSOneRou    &   $\DSTwoCountRound$    &   \DSAllTrainCountRound    \\

    \midrule
    
    \multicolumn{1}{c|}{Instances grouped by pixel area}       &   \multicolumn{4}{r}{}    \\
    \midrule
    \specialcell{Small (< 32x32 px)}                &   \DSZeroSmall    &   \DSOneSmall   &   1455    &   \DSAllSmall    \\
    \specialcell{Medium (32x32 -- 96x96 px)}        &   \DSZeroMed   &   \DSOneMed   &   3345    &   \DSAllMed    \\
    \specialcell{Large (> 96x96 px)}                &   \DSZeroLar   &   \DSOneLar    &   580     &   \DSAllLar    \\
    
	\bottomrule
    \end{tabular}
}
\end{table}

\subsection{Environment setup}
\label{sec:envsetup}


\subsubsection{Hardware}
Object detector training requires a lot of system resources, especially 
with the larger backbones and with increasing dataset sizes. 
We trained our models on a supercomputer cluster that is part of a 
National Super Computing Grid. 
The supercomputer we used in our experiments employs 
682 CPU nodes and its performance can theoretically peak at 
1.8 petaflops. Each node has two 20 core Intel Xeon Cascade Lake processors 
running at 2.1 GHz. 
It also features an ``AI partition'' that includes 80 GPU nodes 
with four Nvidia Tesla V100 32GB GPGPUs each, totaling 320 GPGPUs. 
The total theoretical performance of the GPGPUs is 2.7 petaflops. 
The nodes carry 384 GB of main memory and 3.6 TB quick local storage. 
For our experiments, we used one node that employs four Nvidia Tesla V100 32GB GPGPUs. 
We also performed some intermediate tests on our group's HPC GPU mini-cluster.
It features two Intel Xeon E5-2640 v4 CPUs totaling 20 cores 
running at 2.40 GHz, and includes eight Nvidia Tesla P100 16 GB GPGPUs. 
For our experiments, we used four Nvidia Tesla P100 16GB GPGPUs.

\subsubsection{Software}

In Table~\ref{tab:sw} we present a detailed list of software used during 
our experiments. 
The majority of software in our experiments is based on (or is written in) Python, 
and the frameworks in our experiments were implemented in PyTorch. 
To work with HPC clusters, we also needed to set up a Python environment 
with several needed libraries and packages. In order to achieve the setup, 
a virtual environment is required. Therefore, we used Conda~\cite{conda} 
which is an open-source virtual environment and package manager. 
Furthermore, Conda enabled us to install and use different matching versions 
of packages and libraries, and it also isolated them to the specific 
virtual environment, making the experimentation and failure less painful 
and more streamlined. In particular, we used Miniconda3 variant since we 
did not require the features of the bulkier Anaconda3 package. 

\begin{table}[htb]
\centering
\caption{Software and tools used to perform the experiments.}
\label{tab:sw}
\resizebox{\columnwidth}{!}{%
  \begin{tabular}{lll}
    \toprule
    Software Name & Version & Purpose \\
    \midrule
    Python & 3.8.1 & Python core \\
    cudatoolkit & 10.1.243 & GPU programming \\
    albumentations & 0.4.3 & Image augmentation library \\
	numpy & 1.18.1 & Scientific computing package \\
	OpenCV & 4.2.0.32 & Computer vision library \\
	PyTorch & 1.5.1 & Machine learning framework \\
	torchvision & 0.6.1 & Computer vision package \\
	matplotlib & 3.1.3 & Graph visualizations \\
	pycocotools & 2.0 & Tools for MS COCO \\
	tqdm & 4.42.1 & Progress bar for terminal use \\
	pillow & 7.0.0 & Imaging library (PIL fork) \\
	cython & 0.29.15 & C-Extensions for Python \\
	ninja & 1.9.0 & Small build system \\
	pandas & 1.0.1 & Data analysis library \\
	requests & 2.23.0 & HTTP library \\
	scipy & 1.4.1 & Mathematics and science library \\
	yacs & 0.1.6 & Configurations management system \\
	Tensorboard & 2.1.1 & Training data capture \\
	Detectron2 & 0.1.1 & Object detection framework \\
	mmcv-full & 1.1.4 & OpenMMLab Computer Vision framework \\
	mmdet & 2.2.0 & MMdetection object detection toolbox \\
	mmpycocotools & 12.0.3 & MMdet pycocotools fork \\
	QCC & 8.3.0 & GNU Compiler Collection \\
	CUDA & 10.1.168 & Nvidia CUDA \\
	labelme & 4.2.9 & Annotation tool \\
	ATSS & & ATSS - detector \\
	CenterMask2 & & CenterMask - detector (PyTorch) \\
	TridentNet & & TridentNet - detector \\
	ResNeSt & & ResNeSt - detector \\
	DetectoRS & & DetectoRS - detector \\
	detectron2-pipeline & & Modular image processing pipeline \\
	\bottomrule
  \end{tabular}%
}
\end{table}

\subsection{Model training}
\label{sec:modeltrain}

To date, it took the equivalent of at least \emph{three person-months of effort} 
to perform all the model training experiments, including trials and failures 
as well as validation and testing of intermediate and final models.
It also took the equivalent of at least \emph{600 GPU-hours} of computing effort using the described hardware configurations.

\subsubsection{Initial models}

%
%
%

Our first efforts with \texttt{Dataset0} included three detectors with 
total of six varying models created. We trained 
CenterMask2~\cite{cm_git,lee_park_cm} with VoVNet-V2~\cite{lee_vovnet} 
as backbone using the \emph{V-57-eSE}, \emph{V-99-eSE}, \emph{V-39-eSe (lightweight)} 
variants, ATSS~\cite{zhang_atss}, we used the ResNet-50~\cite{he_resnet} 
and ResNeXt-101~\cite{xie_resxt} backbones with multi-scale training 
and deformable convolutions and TridentNet~\cite{li_trident} using the 
ResNet-101~\cite{he_resnet} C4 backbone. 
As already mentioned, we achieved excellent results with our \texttt{Dataset0}
where the highest precision was for validation set $\BestAccuracyValidPct$
and $\BestAccuracyPct$ for testing set respectively (see Section~\ref{sec:apdx-ds0}). 
Though results were satisfactory, as our model use-cases began to realize 
and given the peer-reviews received, we decided to create a larger dataset 
with more variance in the images. As the field of object detection is 
evolving quickly, we also substituted our detectors for more promising ones.  
As our dataset has been overhauled including a whole new 
testing set and the metrics were modified to suit our needs also, the 
results are not entirely comparable. 
Therefore, we omit further comparison to the older models (\texttt{Dataset0}),
and we focus entirely on our latest developments around \texttt{DatasetAll}.

\subsubsection{Latest models}

Therefore, for \texttt{DatasetAll} we used two state-of-the-art 
object detection frameworks to train our 
models with -- FAIRs detectron2~\cite{d2} and MMdetection~\cite{mmdet} by 
Multimedia Laboratory, CUHK. Our ResNeSt~\cite{resnest} model used 
detectron2~\cite{d2_wu} and DetectoRS~\cite{detectors} model was trained 
on MMDet~\cite{mmdet}. Both of the frameworks are equipped to handle our 
COCO-style dataset by default and feature dataset evaluators 
using the pycocotools library. 

\paragraph{ResNeSt.}

For ResNeSt~\cite{resnest}, we chose a huge 200-layer deep backbone 
with the Cascade R-CNN~\cite{cai_cascade} method. Most of the settings we 
left as standard at this point, therefore, the models feature 
FPN~\cite{lin_fpn}, SyncBN~\cite{ioffe} and image scale variation that 
randomizes the short side of the input image between 640 and 800 pixels. 
We changed the class number to two and configured the training schedule 
in regards to our previous experiments in which we first set the maximum 
iterations to 45000 with (i.e., 0.5x of the ``1x learning rate'' schedule 
which translates to default 90000 iterations). 
However, we found that we get even better results at 20000--40000 iterations 
checkpoints. Therefore, our schedule was set to 36000 maximum iterations 
with lowering steps at 30000 and 34000 iterations. Our base learning rate was 0.02.

\paragraph{DetectoRS.}

For DetectoRS~\cite{detectors}, we chose the offered ResNet-50 based backbone. 
The images were input at 800 pixels short size and the learning rate was set at 0.01. 
The number of classes was set to two and the scheduler was left at 12 epochs. 
With our dataset and the setting at 2 samples per worker resulted 
in 4195 iterations per epoch totaling at 50340 iterations. The learning rate 
was stepped to one-tenth of the previous at the beginning of epoch nine and twelve.

\subsection{Various enhancements}
\label{sec:enhance}

\subsubsection{Image alterations}

We also propose and explore the effect of ``image alteration'' on the 
performance of our trained models~\cite{chen2019image}. 
As ``image alterations'' we applied auto-adjust to 
contrast, equalizer, exposure, and hue-saturation. 
Results improvement with ``image alteration'' vary across the models. 
DetectoRS did not show any signs of improvements on the false detections 
nor the confidence level increase. On the one hand, ResNeSt-model did turn 
a few False Negatives (FN) to True Positive (TP) and False Positives (FP) 
to True Negative (TN) when equalizer and exposure image alterations were applied. 
Still, plenty more tests need to be conducted in order to trust the 
altered images in the ``production models`` rather than the originals.

\section{Analysis of Results}
\label{sec:results}

All the models were tested separately with both the validation dataset
as well as with a separate testing dataset (see Section~\ref{sec:dataset}). 
As mentioned earlier, the testing dataset mainly features images from 
street-level maps, since this corresponds to the intended use for the detector model. 
However, as shown in Section~\ref{sec:apply-reallife} and Appendix~\ref{sec:apdx1}, our system performs 
well (94.6\%) even with arbitrary images acquired by third-parties. 
%

%
Tables~\ref{tab:dsAll-tab1} and~\ref{tab:ds0-tab1} show the training iteration count for each model 
that provided the best performance for the whole training session. 
Additionally, these tables present the timings for training and 
inference under each detection model. 

\begin{table}
\centering
\caption {Final configuration, iterations count, training and inference times, when training detectors on \texttt{DatasetAll}. } 
\label{tab:dsAll-tab1}
\resizebox{\columnwidth}{!}{%
  \begin{tabular}{lcrcrcc}
    \toprule
    Detector & \specialcell{Best-result \\ iterations \\ (count [set])} & \specialcell{Weights \\ file size \\ (MB)} & \specialcell{Avg. train \\ time / iter. \\ (seconds [batch size])} & \specialcell{Avg. inference \\ time / 800px image \\ (seconds)} \\
    \midrule
    \specialcell{ResNeSt} & \specialcell{24999 [test] / 19999 [val]} & 1009 & 1.6 [batch 8] & 0.171 \\
    \specialcell{DetectoRS} & \specialcell{37750 (epoch 9) [test, val] } & 989 & 0.66 [batch 2] & 0.13 \\
	\bottomrule
  \end{tabular}%
}
\end{table}

\subsection{Metrics for evaluation}

%
To evaluate our models, we used \emph{pycocotools} and MS COCO evaluator 
built into the frameworks. For all the models trained, we used the 
same evaluator. The metrics we use and present are modified from the standard MS COCO's~\cite{lin2014microsoft} evaluation metrics to suit our goals. We present the modifications and the arguments for them. We also present a F1-score for our results, a metric derived from the average precision and recall. We measure our F1-score with 0.5 IoU and 100 detections per image thresholds. It gives equal emphasis on both types of false detections. We list the average precision per category (directed and round) as well. These metrics are not part of the standard COCO metrics suite. 
%
To evaluate the performance of a detector for the \emph{detection performance}, 
MS COCO employs 12 characterizing metrics. \emph{Average Precision (AP)} with 
MS COCO represents, in essence, a \emph{mean Average Precision (mAP)} which 
takes the precision average across all the classes, all the while 
\emph{localization accuracy} is built into the precision metrics. 
MS COCO's standard measurement nowadays is largely represented by 
\emph{Average Precision (AP)} and \emph{Average Recall (AR)}, where the 
average is taken on 10 IoU thresholds from 0.5 to 0.95 with a 0.05 interval. 
AP across scales takes into account the area of pixels within 
the segmentation mask or the bounding box. 
In this context, based on the pixel size of the detected segmentation mask 
or the bounding box -- ``small objects'' means areas up to \texttt{32x32} pixels, 
``medium objects'' fit areas between \texttt{32x32} and \texttt{96x96} pixels, 
and ``large objects'' are represented by areas beyond \texttt{96x96} pixels. 

We argue, that for our model use-cases (ie. detection from street view images), 
the ``small objects`` category is not representative or required. We argue that 
considering the varying quality of street view images, the resolution we want 
to run the images at and with computer collected images, objects under 
the \texttt{32x32} pixel threshold are a subject for unnecessary false detections. 
Therefore, we omit the ``small category`` from our results and from further model uses, 
and we prefer to create a more intelligent image capturing algorithm to have 
good coverage over the captured streets. Our datasets still contain instances 
of the ``small category`` and the detection of them can be enabled or disabled at will.
We also drift away from MS COCO with the average recall metrics. We argue that the number of detections past or below a certain reference point is not necessary as we only want the highest possible accuracy from our models. We set the maximum number of detections arbitrarily to 100 as it well past all our instances per image thresholds and should allow the models to perform.

%
%
%
With our main and immediate use-cases, the pixel-precise localization 
is not absolutely necessary, hence the precision with 0.5 IoU 
(i.e., AP@0.5 metric) is most relevant in our case. 
Therefore, we scale all of our results to the 0.5 IoU threshold.
%
%
\subsection{Numerical results}
\begin{table*}[htb]
\centering
\caption {Results for bounding box detection with the \texttt{DatasetAll} \emph{testing set}, 800px short side images, \textbf{bold}=best}
\label{tab:dsA-tab1}
\resizebox{\textwidth}{!}{%
  \begin{tabular}{ccccccccccc}
    \toprule
    Detector & AP@0.5 & AP@0.5:0.95 & APm & APl & AR 100 & ARm & ARl & F1 & \specialcell{AP@0.5 \\ (\emph{directed} type)} & \specialcell{AP@0.5 \\ (\emph{round} type)} \\
    \midrule
    ResNeSt & \textbf{\ResTestEightAP} & \textbf{\ResTestEightAPbig} & \textbf{\ResTestEightAPm} & \textbf{\ResTestEightAPl} & \textbf{\ResTestEightAR} & \textbf{\ResTestEightARm} & \textbf{\ResTestEightARl} & \textbf{\ResTestEightF} & \textbf{\ResTestEightDir} & \textbf{\ResTestEightRou} \\
    DetectoRS & \DetTestEightAP & \DetTestEightAPbig & \DetTestEightAPm & \DetTestEightAPl & \DetTestEightAR & \DetTestEightARm & \DetTestEightARl & \DetTestEightF & \DetTestEightDir & \DetTestEightRou \\
	\bottomrule
\end{tabular}
}
\end{table*}

In Table~\ref{tab:dsA-tab1} we present the metrics (bounding box detection) 
for the ``testing dataset'' with 800-pixel short side images. The results are very good. The decision to cut the tiny samples from the detections is increasing our percentages greatly. Our F1 scores with both of our models are really high, which suggests the lack of false detections in the ``larger detection categories``. Directed samples seem to be harder to detect in our ``testing dataset``. We can also be happy about our localization accuracy as the tough 0.5:0.95 IoU category is in the 70-percentile range. Large samples are clearly no issue for our models and we can certainly suggest that our large sample size is adequate.

\begin{table*}[htb]
\centering
\caption {Results for bounding box detection with the \texttt{DatasetAll} \emph{validation set}, 800px short side images, \textbf{bold}=best}
\label{tab:dsA-tab2}
\resizebox{\textwidth}{!}{%
  \begin{tabular}{ccccccccccc}
    \toprule
    Detector & AP@0.5 & AP@0.5:0.95 & APm & APl & AR 100 & ARm & ARl & F1 & \specialcell{AP@0.5 \\ (\emph{directed} type)} & \specialcell{AP@0.5 \\ (\emph{round} type)} \\
    \midrule
    ResNeSt & \ResValEightAP & \ResValEightAPbig & \ResValEightAPm & \ResValEightAPl & \ResValEightAR & \ResValEightARm & \textbf{\ResValEightARl} & \ResValEightF & \ResValEightDir & \ResValEightRou \\
    DetectoRS & \textbf{\DetValEightAP} & \textbf{\DetValEightAPbig} & \textbf{\DetValEightAPm} & \textbf{\DetValEightAPl} & \textbf{\DetValEightAR} & \textbf{\DetValEightARm} & \DetValEightARl & \textbf{\DetValEightF} & \textbf{\DetValEightDir} & \textbf{\DetValEightRou} \\
	\bottomrule
\end{tabular}
}
\end{table*}

In Table~\ref{tab:dsA-tab2} we show the detection metrics (bounding box detection) for the ``validation dataset''. The results are excellent. The scores are really high for all categories. Cross-referencing the ``testing set`` and ``validation set`` scores, we could argue a slight overfitting issue creeping in, especially with DetectoRS-model. In comparison, the directed subcategory also takes a hit in detection precision. Therefore, we will also argue, that our validation dataset needs yet another revision for improvement in detection generalization. As for now, we are extremely happy with all of the results including the tougher 0.5:0.95 IoU localization tests.
%

\begin{table*}[htb]
\centering
\caption {Results for ResNeSt \emph{segmentation} detection with the \texttt{DatasetAll} \emph{testing and validation sets}, 800px short side images}
\label{tab:dsA-tab3}
\resizebox{\textwidth}{!}{%
  \begin{tabular}{ccccccccccc}
    \toprule
    Set & AP@0.5 & AP@0.5:0.95 & APm & APl & AR 100 & ARm & ARl & F1 & \specialcell{AP@0.5 \\ (\emph{directed} type)} & \specialcell{AP@0.5 \\ (\emph{round} type)} \\
    \midrule
    Test & \ResTestEightAPS & \ResTestEightAPbigS & \ResTestEightAPmS & \ResTestEightAPlS & \ResTestEightARS & \ResTestEightARmS & \ResTestEightARlS & \ResTestEightFS & \ResTestEightDirS & \ResTestEightRouS \\
    Val & \ResValEightAPS & \ResValEightAPbigS & \ResValEightAPmS & \ResValEightAPlS & \ResValEightARS & \ResValEightARmS & \ResValEightARlS & \ResValEightFS & \ResValEightDirS & \ResValEightRouS \\
	\bottomrule
\end{tabular}
}
\end{table*}

In Table~\ref{tab:dsA-tab3} we present the segmentation detection results for the ResNeSt-model. In comparison, we can see that the segmentation results are not far off from the bounding box results. In a few categories, the results are even slightly better. This suggests great localization precision and a great beginning for better identification for camera types as more accurate shapes can be detected as accurately than mere bounding boxes.

Input image resolution is a debatable topic. Should we input images in their native resolution or at the training image size? The ideal case would obviously be if there was no discrepancy between the two. As our testing dataset contains slight higher resolution images than our testing set, we also tested the testing set with 1200 pixel short side input images. We present the results in Table~\ref{tab:dsA-tab4}. 

From the results, we can see that our DetectoRS-model clearly benefits from the increased input resolution achieving the highest scores with the ``testing set``. However, our ResNeSt-model suffers from the increased input resolution and scores lower than with the smaller images. ResNeSt clearly benefits from images that are close to the training resolution. The results imply that testing the models with varying resolution is beneficial and keeping the resolution constant will yield the stable results we are after. As transfer-training these models is not that time-consuming, creating specialized models for each task (different image properties) is most likely beneficial.

\begin{table*}[htb]
\centering
\caption {Results for bounding box detection with the \texttt{DatasetAll} \emph{testing set}, 1200px short side images, \textbf{bold}=best}
\label{tab:dsA-tab4}
\resizebox{\textwidth}{!}{%
  \begin{tabular}{ccccccccccc}
    \toprule
    Detector & AP@0.5 & AP@0.5:0.95 & APm & APl & AR 100 & ARm & ARl & F1 & \specialcell{AP@0.5 \\ (\emph{directed} type)} & \specialcell{AP@0.5 \\ (\emph{round} type)} \\
    \midrule
    ResNeSt & \ResTestTwelveAP & \ResTestTwelveAPbig & \ResTestTwelveAPm & \ResTestTwelveAPl & \ResTestTwelveAR & \ResTestTwelveARm & \textbf{\ResTestTwelveARl} & \ResTestTwelveF & \ResTestTwelveDir & \ResTestTwelveRou \\
    DetectoRS & \textbf{\DetTestTwelveAP} & \textbf{\DetTestTwelveAPbig} & \textbf{\DetTestTwelveAPm} & \textbf{\DetTestTwelveAPl} & \textbf{\DetTestTwelveAR} & \textbf{\DetTestTwelveARm} & \textbf{\DetTestTwelveARl} & \textbf{\DetTestTwelveF} & \textbf{\DetTestTwelveDir} & \textbf{\DetTestTwelveRou} \\
	\bottomrule
\end{tabular}
}
\end{table*}

\subsection{Visual result samples}

To facilitate the understanding of successes, failures, and challenges 
faced by our detectors, we present in this section a selection of relevant 
samples along with some comments. 

\begin{figure}[htb]
  \centering
  \includegraphics[width=\columnwidth]{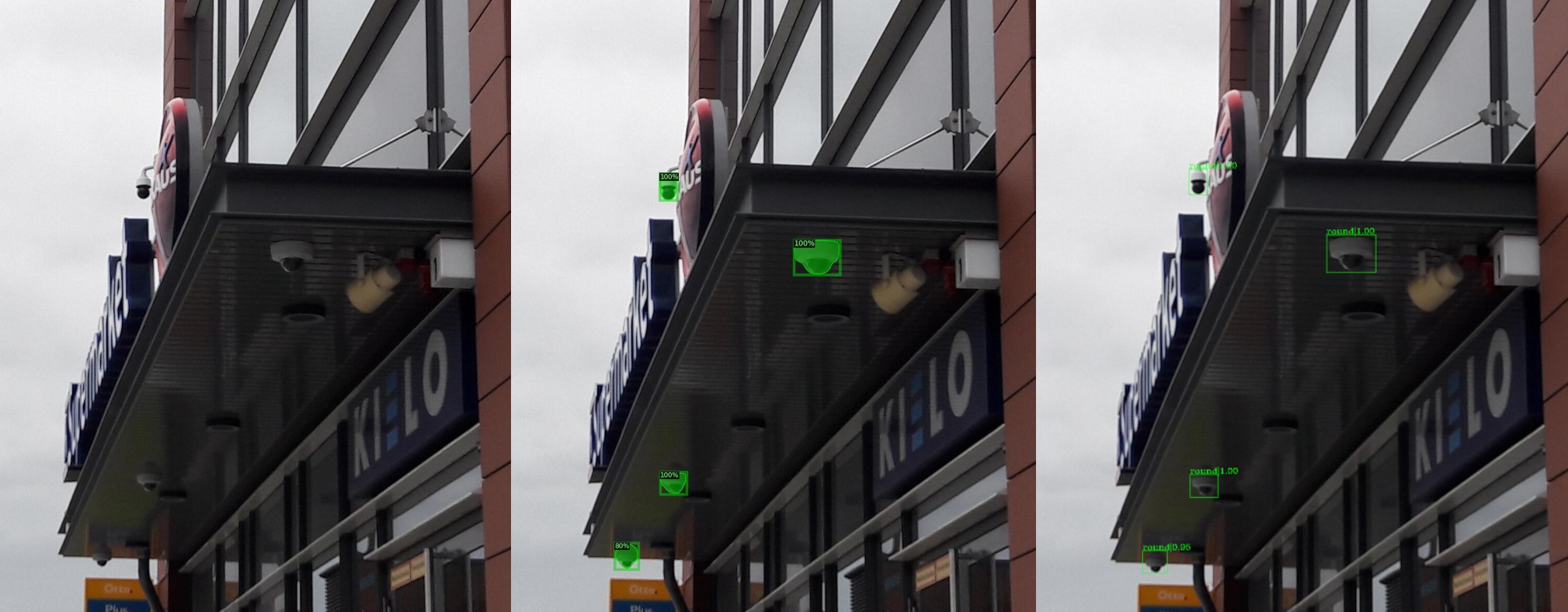}
  \caption{Visual results (Ground Truth - 4 TP) (left to right): ResNeSt - 4 TP (3x100\% and 80\%); DetectoRS - 4 TP (3x100\% and 95\%)}
  \label{fig:img_6}
\end{figure}

\begin{figure}[htb]
  \centering
  \includegraphics[width=\columnwidth]{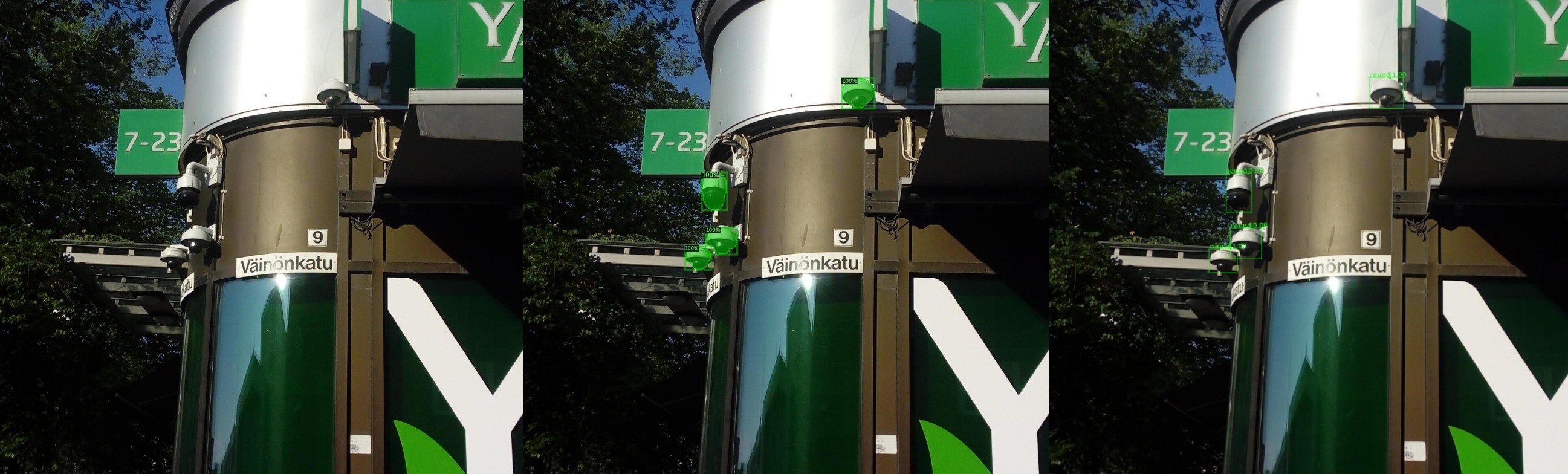}
  \caption{Visual results (Ground Truth - 4 TP) (left to right): ResNeSt - 4 TP (4x100\%); DetectoRS - 4 TP (2x100\%, 99\% and 96\%)}
  \label{fig:img_7}
\end{figure}

\begin{figure}[htb]
  \centering
  \includegraphics[width=\columnwidth]{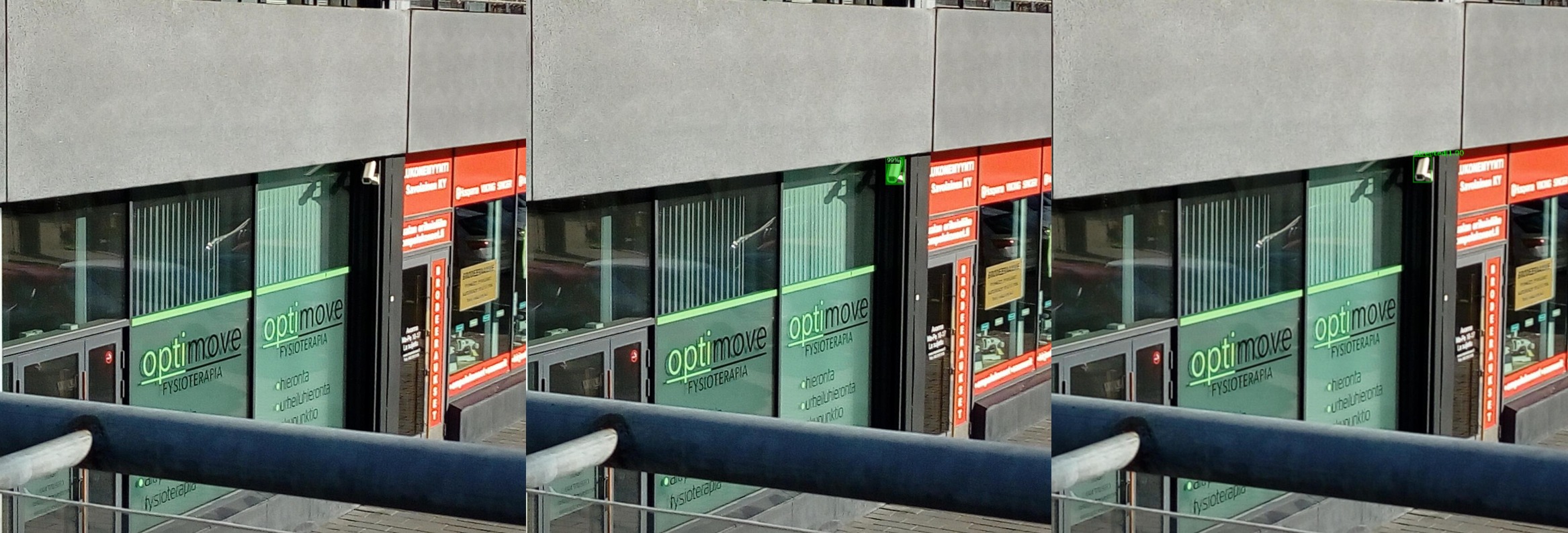}
  \caption{Visual results (Ground Truth - 1 TP) (left to right): ResNeSt - 1 TP 99\%; DetectoRS - 1 TP 100\%}
  \label{fig:img_10}
\end{figure}

Figures~\ref{fig:img_6},~\ref{fig:img_7},~\ref{fig:img_10} are perfect examples 
of our excellent results -- all TPs are found, and nothing else is detected. 
Also the confidence levels on these samples are really high (lowest is 80\%).

\begin{figure}[htb]
  \centering
  \includegraphics[width=\columnwidth]{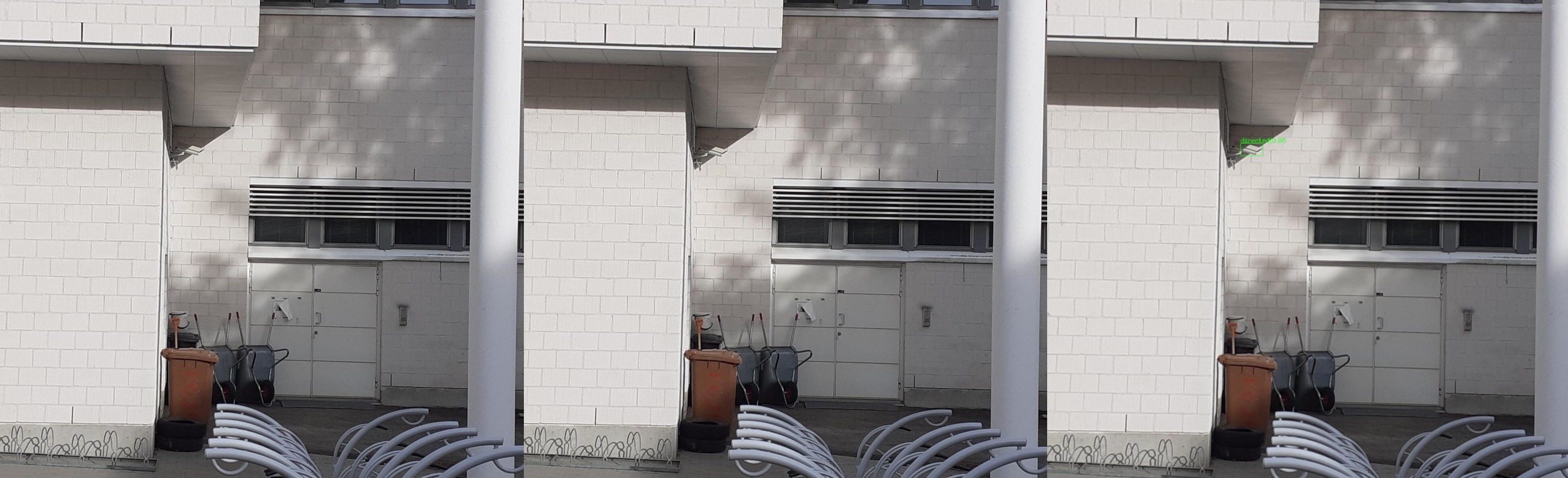}
  \caption{Visual results (Ground Truth - 1 TP) (left to right): ResNeSt - 1 FN; DetectoRS - 1 TP 95\%}
  \label{fig:img_4}
\end{figure}

\begin{figure}[htb]
  \centering
  \includegraphics[width=\columnwidth]{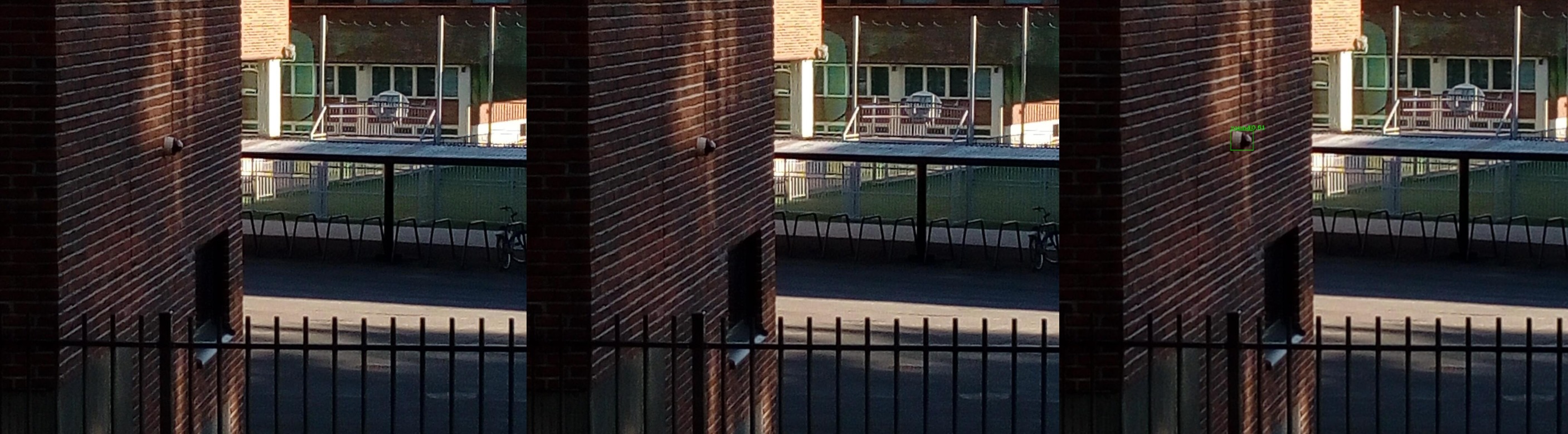}
  \caption{Visual results (Ground Truth - 1 TP) (left to right): ResNeSt - 1 FN; DetectoRS - 1 TP 81\%}
  \label{fig:img_9}
\end{figure}

Figures~\ref{fig:img_4} and ~\ref{fig:img_9} are examples on how 
DetectoRS edges out on some of the samples. In Figure~\ref{fig:img_4}, 
the camera blends into the white wall in the sun and Figure~\ref{fig:img_9} 
is quite dark. ResNeSt-model in unable to detect the cameras, but DetectoRS 
finds those and achieves good confidence in the TPs.

\begin{figure}[htb]
  \centering
  \includegraphics[width=\columnwidth]{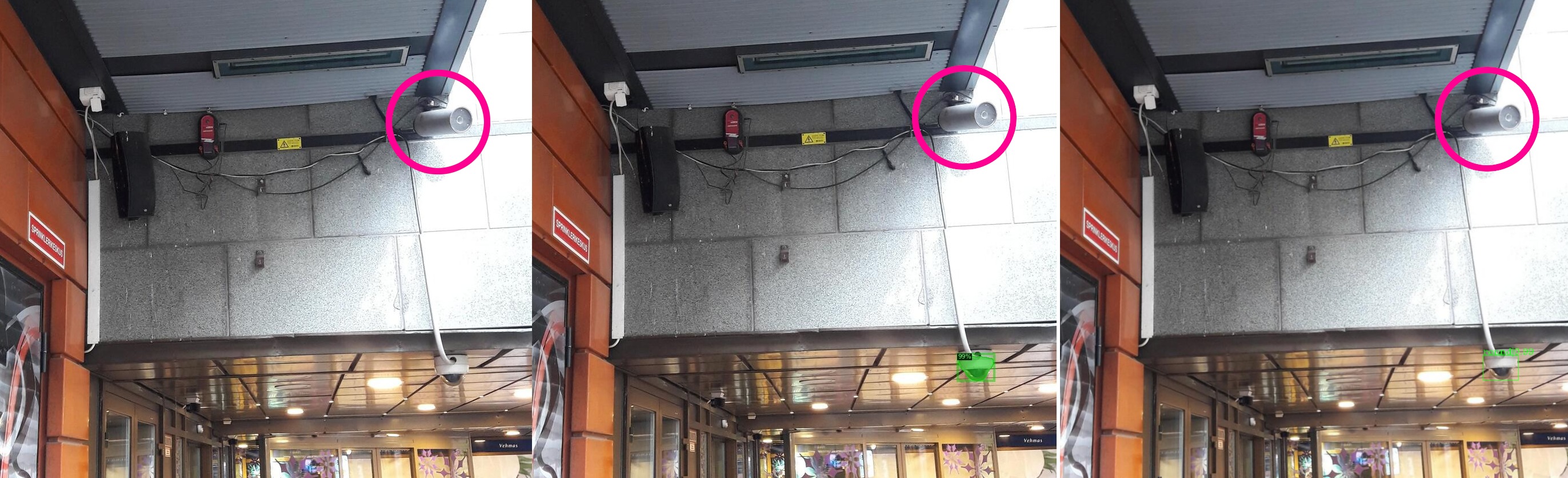}
  \caption{Visual results (Ground Truth - 2 TP) (left to right): ResNeSt - 1 TP 99\%, 1 FN; DetectoRS - 1 TP 99\%, 1 FN}
  \label{fig:img_1}
\end{figure}

Figure~\ref{fig:img_1} showcases a FN on both models. The directed camera 
in the upper corner of the image is undetected. However, the lower round-type 
camera is found in great confidence (99\%).

\begin{figure}[htb]
  \centering
  \includegraphics[width=\columnwidth]{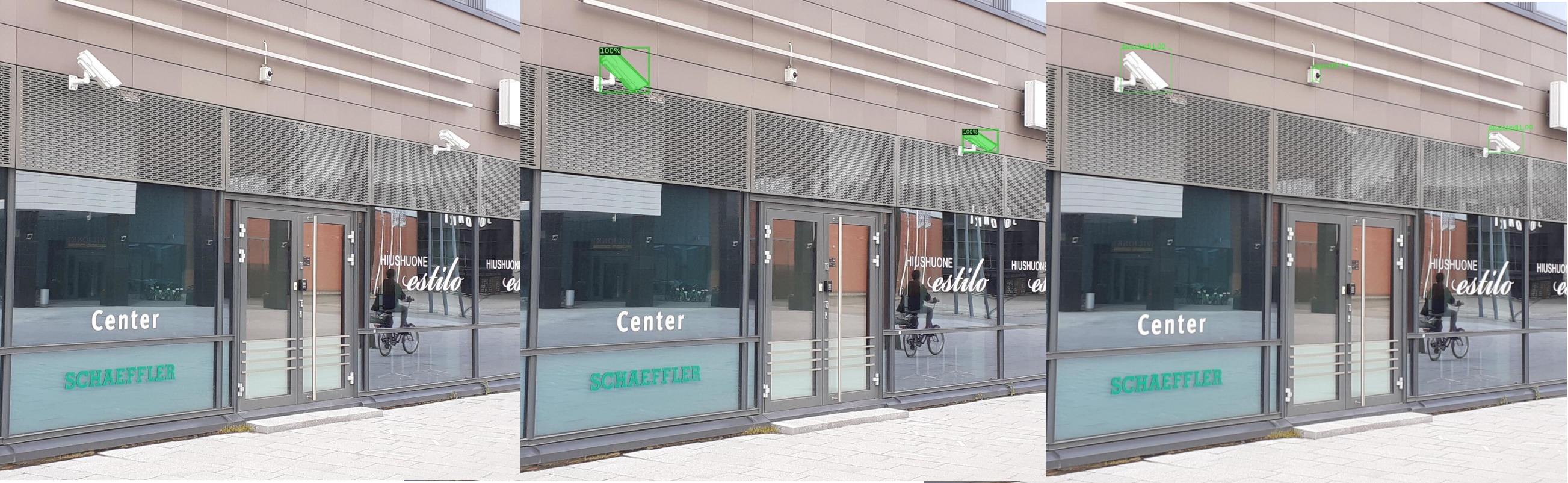}
  \caption{Visual results (Ground Truth - 2 TP) (left to right): ResNeSt - 2 TP (2x100\%); DetectoRS - 2 TP (2x100\%), 1 FP 74\%}
  \label{fig:img_3}
\end{figure}

\begin{figure}[htb]
  \centering
  \includegraphics[width=\columnwidth]{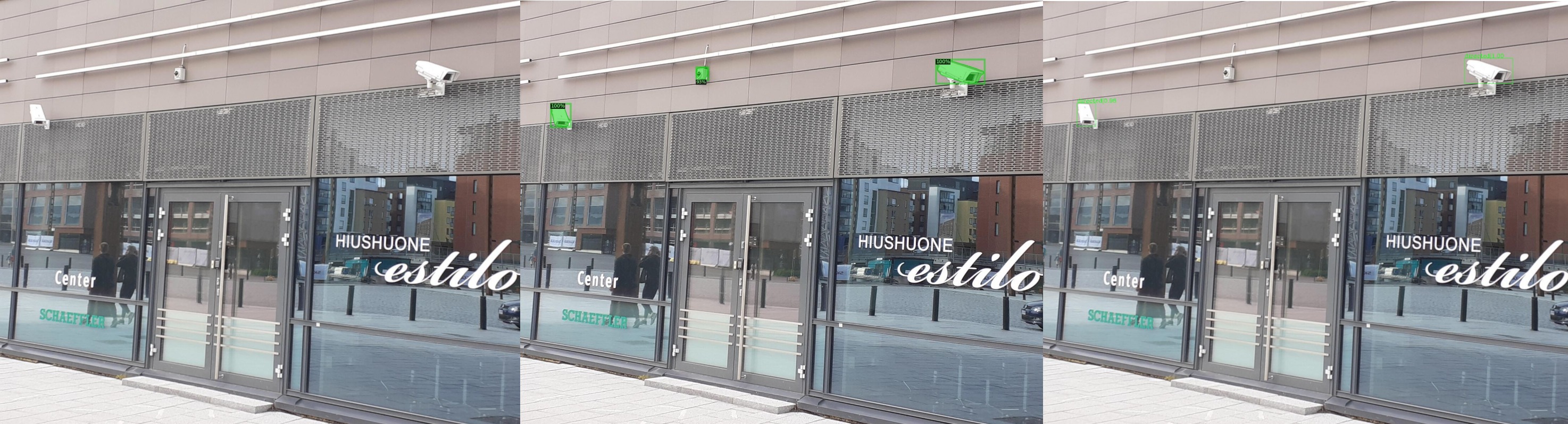}
  \caption{Visual results (Ground Truth - 2 TP) (left to right): ResNeSt - 2 TP (2x100\%), 1 FP 93\%; DetectoRS - 2 TP (100\% and 96\%)}
  \label{fig:img_2}
\end{figure}

Figures~\ref{fig:img_2} and ~\ref{fig:img_3} showcase on how the angle of 
the image can affect the results. Regardless of the angle, both models 
find the TPs, but depending on the angle, both models find a single FP 
on opposite cases. Although, with ResNeSt, the confidence of the FP is 
quite high (93\%) as in the DetectoRS case it is 74\%, which could be 
rooted out with confidence limiting. Sensor fusion could also be used 
here for limiting the possibilities of FPs.

\begin{figure}[htb]
  \centering
  \includegraphics[width=\columnwidth]{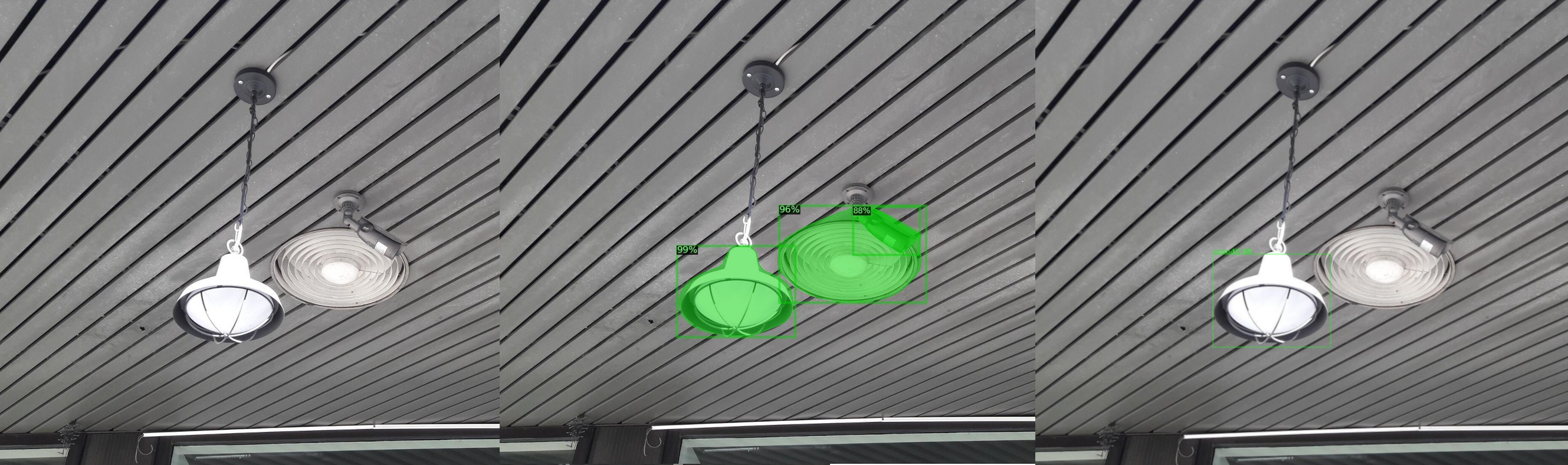}
  \caption{Visual results (Ground Truth - 1 TP) (left to right): ResNeSt - 1 TP 88\%, 2 FP (99\% and 96\%); DetectoRS - 1 FN, 1 FP 95\%}
  \label{fig:img_5}
\end{figure}

Figure~\ref{fig:img_5} is the worst sample on our ``testing set``. 
ResNeSt finds two FPs on the two lamps present, but it also finds the camera. 
DetectoRS only finds a single FP. Confidence levels on the FPs are worryingly high.


\begin{figure}[htb]
  \centering
  \includegraphics[width=\columnwidth]{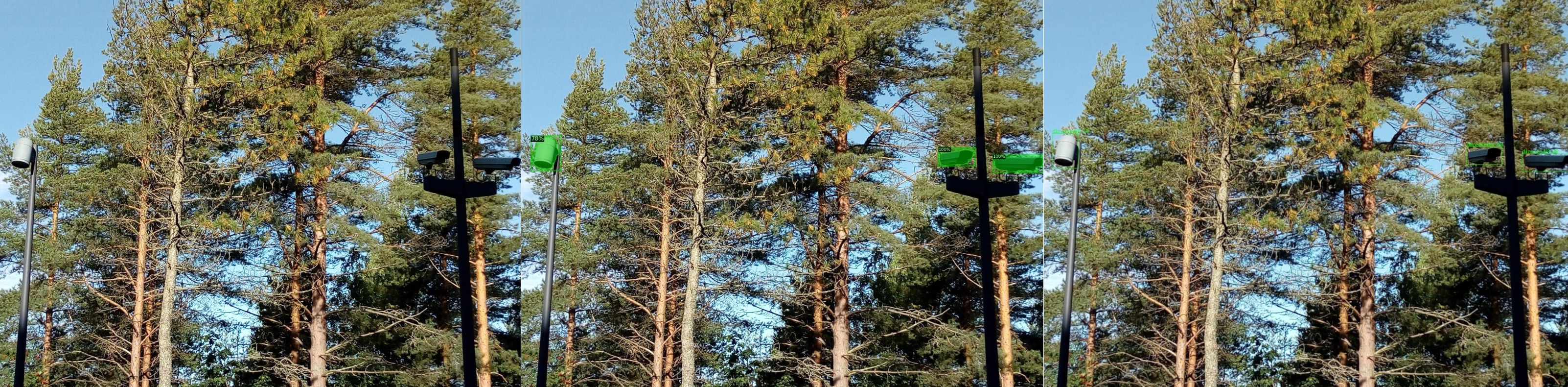}
  \caption{Visual results (Ground Truth - 2 TP) (left to right): ResNeSt - 2 TP (2x100\%), 1 FP 78\%; DetectoRS - 2 TP (100\% and 99\%), 1 FP 73\%}
  \label{fig:img_8}
\end{figure}

In Figure~\ref{fig:img_8}, both models find the TPs easily with high 
confidence. However, both models also find a single FP with 
the lamp. The confidence of those FPs are still lower 
than usual (78\% and 73\%).

Next, we present some examples and results where we applied the 
``image alteration'' techniques (Section~\ref{sec:enhance}), thus seeking 
improvements in confidence levels and TP/TN/FP values that are as close as
possible to the Ground Truth (GT). 

\begin{figure}[htb]
  \centering
  \includegraphics[width=\columnwidth]{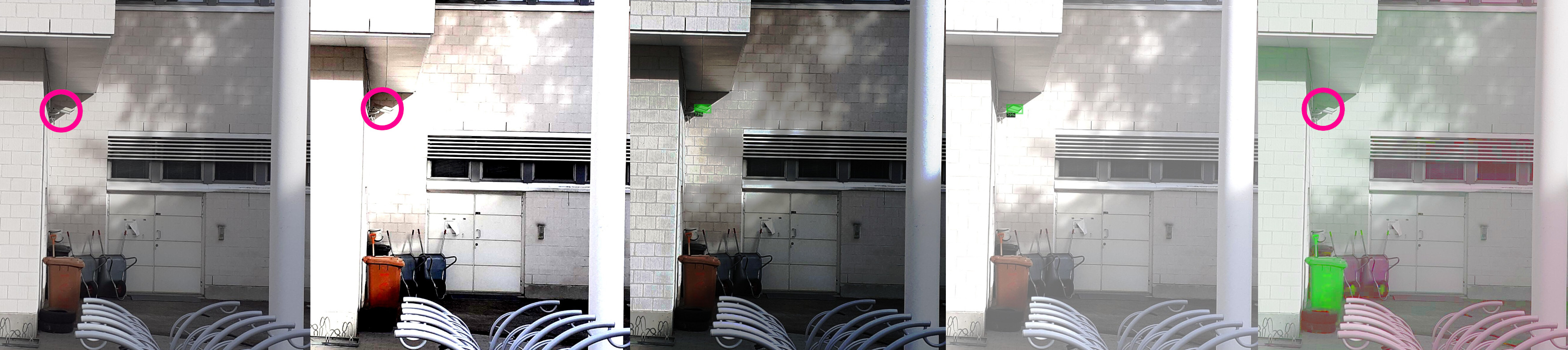}
  \caption{Visual results (Ground Truth - 1 TP) (left to right): Original - 1 FN; Contrast - 1 FN; Equalizer - 1 TP 73\%; Exposure - 1 TP 79\%; Saturation - 1 FN}
  \label{fig:res_alter_2}
\end{figure}

\begin{figure}[htb]
  \centering
  \includegraphics[width=\columnwidth]{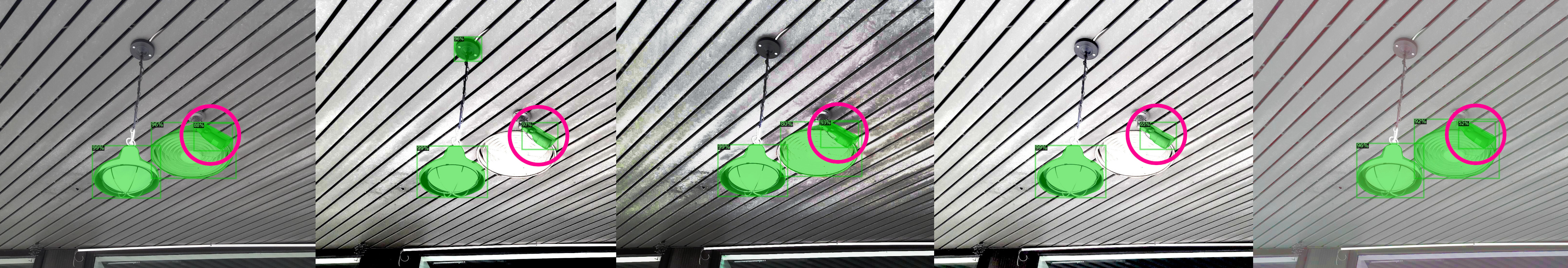}
  \caption{Visual results (Ground Truth - 1 TP) (left to right): Original - 1 TP 88\%, 2 FP (99\% and 96\%); Contrast - 1 TP 97\%, 2 FP (99\% and 96\%); Equalizer - 1 TP 93\%, 2 FP (99\% and 80\%); Exposure - 1 TP 65\%, 1 FP 99\%; Saturation - 1 TP 52\%, 2 FP (96\% and 92\%)}
  \label{fig:res_alter_3}
\end{figure}
  
Figures~\ref{fig:res_alter_2} and ~\ref{fig:res_alter_3} showcase the result 
of ``image alterations'' on ResNeSt-model. In Figure~\ref{fig:res_alter_2}, 
we find improvements as a FN is turned into a TP in exposure and 
equalizer tests. In Figure~\ref{fig:res_alter_3}, the results are a mixed bag. 
On the exposure test, we remove a single FP, but the confidence level 
of the TP camera is hindered and still a single FP is present.

\section{Discussion}
\label{sec:discuss}


\subsection{Practical applications}

\subsubsection{Fast and accurate CCTV camera annotations}
\label{sec:annot}
Crowd-sourcing is a proven and effective method for fast, accurate, and 
cost-effective for image labeling and annotation, and generic 
computer vision and machine learning tasks~\cite{kovashka2016crowdsourcing,wah2006crowdsourcing,welinder2010online,di2013crowdsourcing,deng2013fine}. 
Google is using a similar approach integrated into its 
reCAPTCHA V2~\cite{von2008recaptcha,recaptcha_help}.
In Figure~\ref{fig:idea-recaptchav2}, we show our vision to 
extend the current reCAPTCHA V2 system. 
Our proposed improvement could ask the users of reCAPTCHA to 
\emph{``Select all images with \textbf{CCTV cameras}''}, 
hence leveraging reCAPTCHA's unified infrastructure and algorithms 
to better and faster help to annotate and validate the CCTV camera objects 
in Google Street View imagery as well as other image datasets. 
To our knowledge, at the time of this writing there is no such publicly 
available feature in Google's reCAPTCHA.

\subsubsection{Global and instant mapping of CCTV locations and areas}
\label{sec:apply-mapping}

On the one hand, at present there are multiple projects and data-sources that 
provide the geo-location mapping of CCTV cameras.
Some of these projects are open-source and crowd-sourced~\cite{cctv-anopticon,cctv-osm}, 
while some others are open-data resources provided by city administrations and 
country governments~\cite{cctv-moscow,cctv-paris1,cctv-paris2,cctv-brussels}.
However, all of these feature a set of major limitations, as follows. 
First, all these approaches cover a very limited geographical area 
(e.g., maximum a city). 
Second, the data-sources and projects are globally uncoordinated. 
Therefore the data formats along with the exposed characteristics of 
the respective maps and CCTV cameras vary dramatically across the board.
Third, the crowd-sourced project relies heavily on human contributions of data, 
while governmental data-sources rely on human administration of data. 
Such an approach is highly unscalable for maintenance and development of 
the datasets, and exposes the data to human error and (un)intentional 
manipulations. 
Fourth, those datasets are rarely kept up-to-date and they inherently 
cannot reflect the instant changes in the addition, removal, reposition of
CCTV cameras (e.g., due to infrastructure changes, construction).

On the other hand, our CCTV camera object detector can be applied 
to a global source of street-level imagery such as Google Street View, 
OpenStreetCam and, Mapillary. This allows fast mapping and localization 
of most street-level (and even indoor~\footnote{Our approach can also detect 
and map indoor CCTV cameras thanks to panoramas in Google Street View 
supplied by Google and its users~\cite{gsv-pano}.}) 
CCTV cameras, therefore instantly creating and maintaining a global 
up-to-date (and historical~\cite{gsv-backtime}) map of CCTV cameras. 
At the same time, our CCTV camera object detector can be used to validate 
the accuracy and truthfulness of the crowd-sourced and open-data datasets. 
To achieve this, an automated process picks each CCTV camera entry from a
dataset, retrieves the relevant and closest street-level and geo-tagged 
imagery, applies our CCTV camera object detector, and finally validates 
if the dataset is correct and contains up-to-date information.

\subsubsection{CCTV-aware route planning and navigation}

Once the CCTV cameras can be accurately detected, and instantly 
located and mapped based on street-level imagery~\footnote{Also can be applied to geotagged photos and videos.} 
(see Section~\ref{sec:apply-mapping}), the system is ready for one of 
the most important and relevant application -- \emph{CCTV-aware route planning and navigation}.

At present, there is a myriad of route planning algorithms, software, and 
services~\footnote{A wide variety of open-/closed-source, 
online/offline, free/paid solutions.}~\cite{osm-route-online,osm-route-offline,luxen2011real,delling2009engineering,bast2016route,szczerba2000robust}.
However, to the best of our knowledge, none of the currently available 
algorithms, software, and services provide \emph{CCTV-aware} route planning and navigation.
To the best of our knowledge, we are the first to propose and work on such 
features at present and we are unaware of any project or service developing 
or offering such route planning options.

\begin{figure}[htb]
    \centering
    \includegraphics[width=1.0\columnwidth]{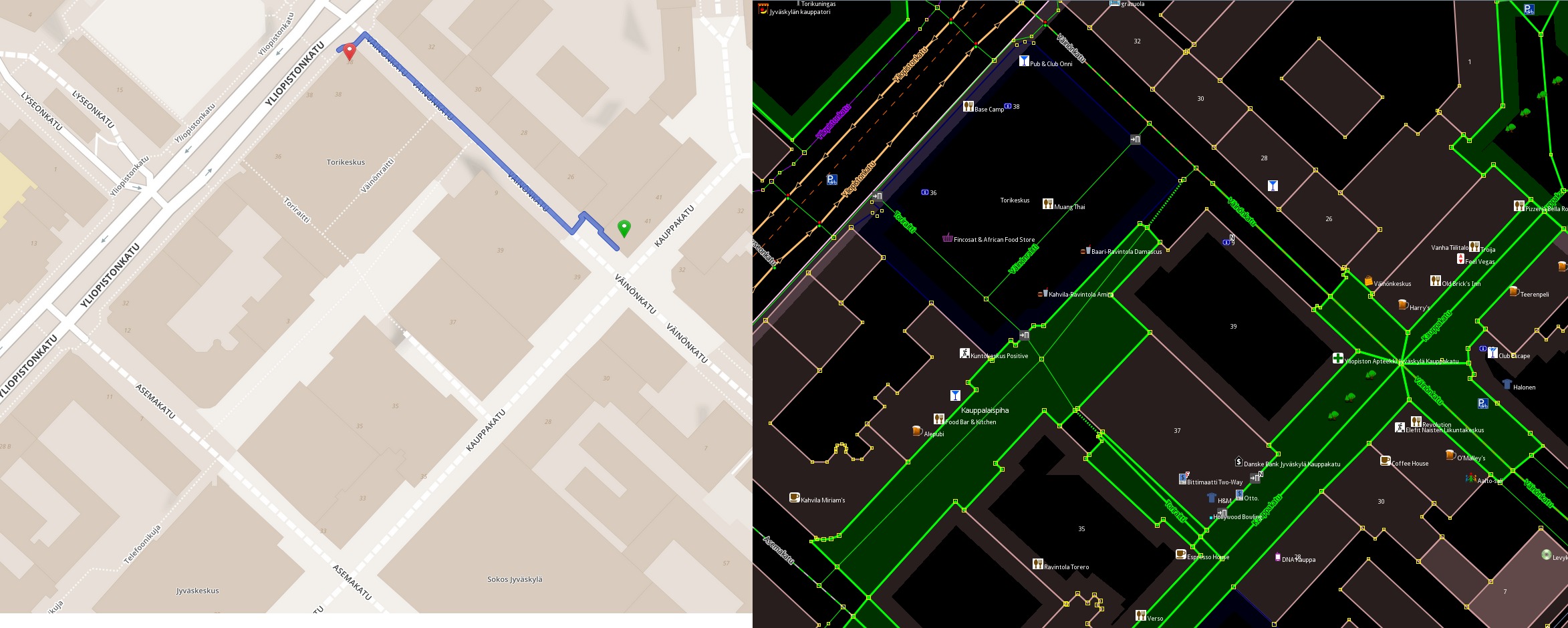}
    \caption{Excerpt from our \textbf{working prototype}: 
    default route is shown as obtained with OSRM as a core routing engine -- CCTV-aware customization and cameras locations \emph{disabled}. 
    }
    \label{fig:routes-bad-osm}
\end{figure}

\begin{figure}[htb]
    \centering
    \includegraphics[width=1.0\columnwidth]{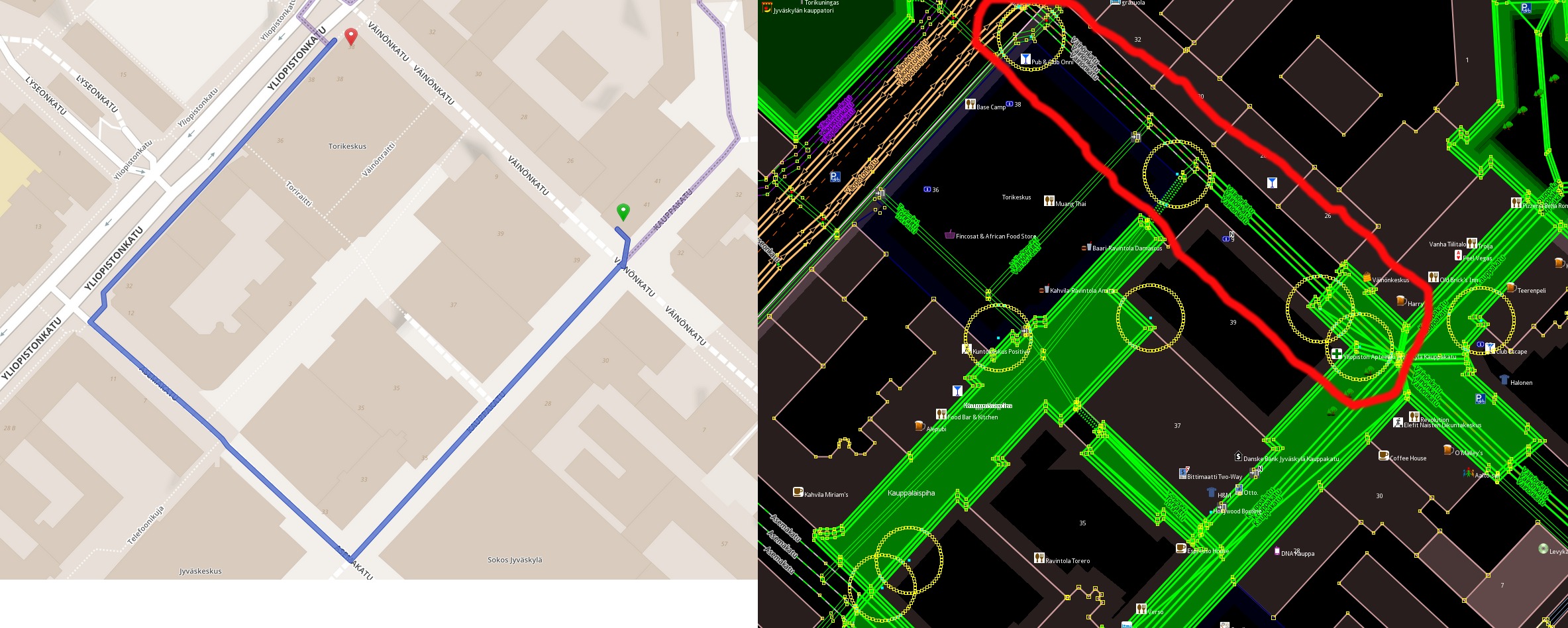}
    \caption{Excerpt from our \textbf{working prototype}: 
    CCTV-aware routing system providing route for 
    \emph{``Avoid CCTV Cameras (privacy-first)''} user option 
    -- CCTV-aware customization and cameras locations \emph{enabled}. 
    }
    \label{fig:routes-good-osm}
\end{figure}

In Figures~\ref{fig:routes-bad-osm},~\ref{fig:routes-good-osm},~\ref{fig:routes-bad-gmaps},~\ref{fig:routes-good-gmaps} 
we demonstrate some real-world use-cases of such a CCTV-aware route planner.
The samples contain the mapping and labeling of location, 
type (e.g., \emph{round camera}, \emph{directed camera}), and field of view. 
This mapping and labeling is achieved using the previously detailed 
CCTV camera object detectors (Sections~\ref{sec:method},~\ref{sec:results}). 
The user first have the option to select the type of \emph{CCTV-aware routing}, 
as presented for example in Figure~\ref{fig:routes-gmaps}. 

If the user selects the \emph{``Follow CCTV Cameras (safety-first)''} option, 
the system would provide a route as in Figures~\ref{fig:routes-bad-osm},~\ref{fig:routes-bad-gmaps}. 
As already mentioned, current route planning solutions provide 
\emph{CCTV-unaware} algorithms, therefore the systems we tested 
such as Google Maps, OpenStreetMap (OSM), by default provided almost always 
non-privacy-preserving routes (e.g., Figures~\ref{fig:routes-bad-osm},~\ref{fig:routes-bad-gmaps}). 

At the same time, with our \emph{CCTV-aware} route planning proposal, the 
user may alternatively select \emph{``Avoid CCTV Cameras (privacy-first)''}. 
Therefore, our system would provide a better route for that 
scenario, as shown in Figures~\ref{fig:routes-good-osm},~\ref{fig:routes-good-gmaps}. 

Below we shortly explain the main excerpts taken from working prototypes. 
In Figure~\ref{fig:routes-good-osm} (right), we enabled the CCTV-aware 
customizations and placed some ``round cameras'' on the map representing 
some real-life cameras and their actual positions as collected and mapped 
in a city that is covered by our system. 
Compared to Figure~\ref{fig:routes-bad-osm} (right), there are now at least 
four ``round cameras'' between starting and ending route points 
(highlighted within the red area) as shown in Figure~\ref{fig:routes-good-osm} (right). 
Given this configuration, we enabled the CCTV-aware customizations 
in our prototype Open Source Routing Machine (OSRM) backend~\cite{osrm}. 
As expected, in Figure~\ref{fig:routes-good-osm} (left) 
it produced an acceptable \emph{privacy-first} routing that is different from 
the default OSM/OSRM routing in Figure~\ref{fig:routes-bad-osm} (left). 
Also as expected, the \emph{privacy-first} routes may not necessarily be optimal 
in terms of travel distance or time. Indeed, increased travel time and dinstance 
is one of the trade-offs a person may need to accept in order to benefit 
from \emph{privacy-first} routing. 

At a very high-level, our \textbf{working prototype} of CCTV-aware map 
routing and navigation uses the following components:
\begin{itemize}
\itemsep0em
\item \textbf{OSM}: access to global mapping and street data, and programatic APIs to access and manipulate its data
\item \textbf{OSRM backend}: efficient route calculations based on various constraints -- we model CCTV cameras' areas of coverage as certain type of constraints in OSRM
\item \textbf{OSRM frontend}: UI/UX and testing/validation of routing and navigation correctness
\item \textbf{JOSM editor}: visualization, editing, and debugging of routing and CCTV data
\item \textbf{Street widths data}: location-specific data -- this may affect CCTV coverage over a particular route
\end{itemize}
Due to space constraints and the depth of the topic, we leave the detailed 
presentation and evaluation of our 
\emph{CCTV-aware privacy-/safety-first route planning and navigation} 
for one of the upcoming publications.

\subsubsection{Real-time CCTV detection on mobile/IoT devices}
\label{sec:apply-mobile}

Finally, one more application of the presented object detector is 
its use for real-time CCTV camera detection on smartphones, mobile 
equipment (e.g., robots, drones), and other IoT/edge devices 
(e.g., RaspberryPi, ESP32-cam, Nvidia Jetson). 
For this purpose, our object detectors are configured and trained for 
low-power devices using, for example, TinyML hardware with TensorFlow 
Lite~\cite{banbury2020benchmarking,zhang2018pcamp}, 
or variants of the state-of-the-art 
MobileNet~\cite{howard_mob,sandler2018mobilenetv2,mobilenetv2_github}, 
and GhostNet~\cite{han2020ghostnet,ghostnet_github}. 

One use-case could be drones or mobile robots equipped with tiny/lite versions of our 
detection models could move around the corner or down the street to first detect 
and map the CCTV areas (i.e., data collectors). 
%
%
The device would then alert the user 
when it detects in real-time a CCTV camera in its field of view, therefore 
providing the user a chance to change course and make a more informed decision. 

At present, we are experimenting with several variants of \emph{CCTV-aware mobile devices}, 
such as the ones based on RaspberryPi4 (RPi4) and Nvidia Jetson Nano.
Figure~\ref{fig:rpi4_pantilt_coral} presents one of our RPi4 
prototypes that we built for our complete CCTV-aware experimental set. 
Our RPi4 prototype is equipped as follows. 
One PiCamera (mounted on the Pan-Tilt HAT) provides both image/video 
acquisition (e.g., up to real-time 24-30 FPS if needed). 
One Coral USB provides AI/ML acceleration to enable real-time object 
detector loaded and running as lightweight model within Google's Coral chipset. 
One GPS-dongle is for accurate localization of the device (e.g., 5--10 meters 
with GPS II chipsets). 
One programmable UART/TTL laser-based range finder (mounted on the Pan-Tilt HAT) 
enables very accurate ($\pm$ 2mm error at 60m range) distance measurement to the detected CCTV cameras. 
One Pan-Tilt HAT enables PiCamera movement for centering and 
focusing the PiCamera street imagery or on detected CCTV cameras, as well 
as the movement of the UART/TTL laser-based range finder. 
At a very high level, the principle of operation of one such device 
mounted on backpack, helmet, bicycle, car rooftop/windshield, 
is presented in Listing~\ref{lst:mobiledev}. 
In our experiments, the maximum drawn power consumption for the entire prototype was 
$7 W$, i.e., $1,4 A$ at $5 V$ RPi4 supply. For example, using a standard yet high-performance 
power-bank of $30000 mAh$, it could operate around 20--24 hours. 

Due to space constraints and the depth of the topic, we leave the detailed 
presentation and evaluation of our 
\emph{mobile device CCTV-aware camera detectors and data collectors} 
for one of the upcoming publications.

\subsubsection{Validation on third-party data in real-life}
\label{sec:apply-reallife}

In this context, we also tested our system 
(namely the ATSS X-101 and Trident R-101 models) 
against the real-world journalist experiment by Pasley~\cite{bi2019cctv}, 
where the ``ground truth'' consists of 39 TPs (i.e., CCTV cameras). 
Running ATSS X-101 detector resulted in 33 TPs + 0 FPs + 6 FNs. 
Applying Trident R-101 model produced 33 TPs + 1 FPs + 6 FNs. 
Both models presented an F1-score of 91.7\%, and while their TPs and 
FNs counts are the same, the actual distribution of TPs and FNs across 
the input images slightly differ. 
Therefore, we also evaluated our system using a ``sensor fusion'' approach~\cite{gustafsson}, 
where the best results from both the detectors (i.e., ``sensors'') are 
combined (i.e., ``fused'') together for an enhanced final result.
%
In this case, our system achieves 35 TPs + 0 FPs + 4 FNs with an F1-score of 94.6\%. 
A visual results subset from our experiment can be seen in Appendix~\ref{sec:apdx1}.
This, therefore, underpins once again that our technology could be 
useful as \emph{privacy-first} early warning system against CCTV cameras for 
real-life use by third parties and users.

\subsection{Technical challenges}

Based on our real-world and street imagery observations, light equipment 
(e.g., light poles, light fixtures) many time look very similar to 
CCTV cameras. This makes it challenging to certainly distinguish between 
CCTV cameras and light equipment even to an experienced human observer. 
The worst-performing sample in our ``testing set`` is Figure~\ref{fig:img_5} 
where the FPs are in fact round lamps or light fixtures. 
It is a perfect example that such a scenario poses certaing challenges. 
Most of the FPs resulted where ``small`` size TPs were present.
However, FPs resulted also for a small number of TPs of ``medium'' and ``large'' sizes, 
for example shapes on walls where the light hits the wall just at specific angle. 
These types of FPs will happen, and the way to root them out is possibly 
to use more intelligent image capturing and some sensor fusion techniques.
At the same time, there is a growing trend to have streetlights 
with embedded CCTV cameras~\cite{sandiego2020lights}. We suspect 
this may pose additional challenges (not insurmountable though) 
for the most accurate detection of CCTV cameras in real-life scenarios.

\subsection{Ethical aspects and potential abuse}
\label{sec:ethic}

On the one hand, there are less technological domains where a 
system inspired by our work could be applied.
For example, the case of Novichok (A-234) poisoning in Salisbury (UK) of the former 
Russian military intelligence officer Sergei Skripal, and his daughter 
Yulia Skripal, by what is believed to be two officers of Russian GRU~\cite{team2018skripal,team2018full}, 
is surreal and beyond infamous~\cite{farrell2020assassination}.
Following the attack and the international diplomatic fallout featuring 
sanctions and diplomatic expulsions, the Law Enforcement agencies and 
investigative think-tank organizations (e.g., Bellingcat) reconstructed 
the complete routes with exact geo-positions, timestamps, and visual proof 
of main suspects \emph{with the heavy use of state- and privately-owned CCTV cameras}~\cite{guardian2018novichok}.
In this context, some readers may object that future potential delinquents 
could use a system similar to ours in order to map the CCTV cameras within 
a planned operation area (if not already available), and then use it to 
carefully generate a specific route that provides close to 100\% anonymity. 
We argue our point of view on this at the end of this section. 

On the other hand, there are offensive cybersecurity scenarios 
(possible but unlikely in the immediate future), 
where our technology could be potentially used. 

\textbf{Attack description:}
The scenario works under the following attack assumptions:
\begin{itemize}
\itemsep0em
\item it is completely autonomous (i.e., \emph{no human operators})
\item it involves offensive drones equipped with CCTV camera object detector (\emph{attackers}) (see Section~\ref{sec:apply-mobile})
\item it targets highly-valuable air-gapped network(s) to exfiltrate data from (\emph{victims})
\end{itemize}
The drones use the object detector to identify CCTV cameras and approach them. 
Once approached, the drones would send a ``backdoor knock'' signal (e.g., visual, 
sound, infra-red) to the camera~\cite{costin2016security,guri2019air}. 
If a particular CCTV camera was internally compromised and connected to 
an isolated network segment that contains a latent malware implant 
(e.g., APT), the malware activates when the camera receives the 
``backdoor knock'' signal. 
The malware could also be implanted in the CCTV camera 
itself via IoT malware and firmware modifications~\cite{cui2013firmware,antonakakis2017understanding,costin2018iot}. 
To complete the attack, the drones with such CCTV camera object detection 
and data-exfiltration capabilities follow the \emph{air-jumper} attack 
described by Guri and Bykhovsky~\cite{guri2019air}.

All in all, we argue that while our CCTV camera object detector is 
designed to be used for \emph{positive impact}, the potential of our proposed 
system being misused is similar and comparable to any other 
system (e.g., Kali Linux, Metasploit), 
method (e.g., penetration testing, reverse engineering), 
or device (e.g., kitchen knives in supermarkets). 
Additionally, we argue that the benefits of our system for the majority of 
\emph{positive impact} applications outweigh the risks of misuse for a fraction of 
\emph{negative impact} applications (where we believe the perpetrators 
to be able to find other ingenious ways for illegal or unethical activity, 
should a system like ours not exist). 



\section{Related work and state of the art}
\label{sec:relwork}
\subsection{Object detection systems}
\subsubsection{Detectron2}

Detectron2~\cite{d2} is \emph{``Facebook AI Research (FAIR)'s next generation 
software system that implements state-of-the-art object detection algorithms''}. 
It is a complete rewrite of the open source object detection platform 
Detectron~\cite{girshick2018detectron}, and originated from FAIR's 
successful maskrcnn-benchmark project~\cite{mrcnn}. 
FAIR suggests that the modularity of the framework grants the users 
flexibility and extensibility to train and use state-of-the-art 
object detection algorithms with various system configurations ranging 
from single GPU PCs to multi-GPU clusters~\cite{d2}.

\subsubsection{ResNeSt}
ResNeSt - Split-Attention Networks was introduced in April 2019 by H. Zhang et al.~\cite{resnest}. It features a novel building block to improve the standard ResNet backbone structure, which was originally developed for image classification rather than object detection. The authors introduce their modular Split-Attention block to enhance the performance of the ResNet architecture and gear it towards more downstream object detection and segmentation tasks. Their ResNeSt network consists of multiple Split-Attention blocks stacked to the ResNet structure. In general, each Split-Attention block divides the feature maps into subgroups. The feature representation of the group is derived from a weighted combination of even smaller feature divides called splits. \cite{resnest}. Zhang et al.~\cite{resnest} showcase their state-of-the-art results using the ResNeSt backbone on Cascade RCNN~\cite{cai_cascade} -model. 

%
\subsubsection{MMDetection}

MMDetection is a rich object detection toolbox created by K. Chen et al.~\cite{mmdet}. It is a highly efficient and modular state-of-the-art toolset featuring lots of frameworks and it is running on the back of PyTorch. It features object detection and instance segmentation tools and authors claim state-of-the-art training speeds. It also won the 2018 COCO Challenge object detection section. \cite{mmdet}. Their paper submission for the public was in June 2019 and currently MMDetection is in version 2.

\subsubsection{DetectoRS}
DetectoRS paper was released in June 2020 by S. Qiao et al.~\cite{detectors}. 
It features two distinct mechanisms to improve on object detection tasks. 
They propose Recursive Feature Pyramid (RFP) on top of regular 
FPN~\cite{lin_fpn} structure. RFP features increased feedback connections 
from the FPN~\cite{lin_fpn} to the bottom-up backbone layers in turn 
adopting ``looking and thinking twice`` design. The authors explain that 
this creates more powerful feature representations~\cite{detectors}.
Another distinct feature of the DetectoRS is the Switchable Atrous Convolution (SAC). 
It enables the backbone to conditionally adapt to different scales of objects. 
The gist of SAC is that it doesn't require retraining of previous weights as 
it can adapt the standard convolutions to conditional convolutions 
effectively~\cite{detectors}. 
To date, DetectoRS using the ResNet-50 and ResNeXt-101-32x4d backbones achieves 
state-of-the-art results in COCO test-dev tasks~\cite{detectors}.

\subsection{Object detection datasets}

Lin et al. \cite{lin2014microsoft} proposed a novel dataset for general object 
detection called Microsoft's Common Objects in COntext (COCO, or MS COCO). 
The most recent (2017) update of MS COCO has a fully annotated training dataset 
containing 118000 images and a 5000 image validation dataset. 
In addition, a 41000 image testing dataset is also available. 
There are more than 80 different classes for state-of-the-art object detector 
testing with the MS COCO~\cite{coco_home}, including pedestrians, traffic 
lights, cars, and even teddy-bears.
However, it does not contain object detection models for CCTV cameras, 
though they are currently an indispensable part of any street and 
city infrastructure such as traffic lights and road signs. 

PASCAL Visual Object Classes (VOC)~\cite{eve_voc} project ran object 
detection challenges between 2005--2012~\cite{voc_home}. 
For the 2012 challenge, PASCAL VOC dataset labeled 20 classes 
with both training and validation sets, and provided more than 11000 images 
totaling over 27000 instances~\cite{eve_devkit}. 
The dataset contained several similar classes to increase 
difficulty~\cite{Jiao}. 

ImageNet Large Scale Visual Recognition Challenge (ILSVRC)~\cite{russakovsky} 
is an ongoing annual object category classification and detection challenge 
that has been run since 2010. Russakovsky et al.~\cite{russakovsky} envisioned 
ILSVRC to follow PASCAL VOC~\cite{eve_voc} footsteps in providing a 
challenging dataset and hosting competitions. ImageNet provides about 
1.2 million images for training and around 150000 images for validation 
and testing. ImageNet has 1000 classification labels, 200 being 
originally chosen for object detection challenges~\cite{russakovsky}. 

Open Images Dataset V6 contains over 9 million annotated images for a total 
of more than 15 million bounding boxes covering about 600 object 
classes~\cite{OpenImages}. The project holds the annual 
Robust Vision Challenges, and its goal is to further improve the 
development of computer vision systems~\cite{OpenImagesHome}.

In this context, our work is novel and perfectly extends the state of the art 
by contributing both the methods (e.g., code, scripts), as well as annotated data 
and trained models necessary for object detection of two classes 
of CCTV cameras (directed and round -- see Section~\ref{sec:dataset}).

\subsection{Computer vision / object detection on street-level imagery}

There are numerous companies and projects capturing and serving street-level 
imagery such as Google Street View, EveryScape, Mapjack, 
Microsoft StreetSide, Yandex Street Panoramas, OpenStreetCam, Mappilary. 
Such imagery allows much richer online and offline experiences boosted by 
technological advances in Computer Vision (CV), Machine Learning (ML), 
Natural Language Processing (NLP) with text mining, 
Virtual/Augmented/eXtended Reality (VR, AR, XR).
Overall, capturing street-level imagery presents both tremendous challenges 
and opportunities~\cite{anguelov2010google}.

Paiva~\cite{paiva2018inferring} used \textit{"computer vision to infer urban 
indicators on google street view"}.
Wojna et al.\cite{wojna2017attention} presented a method for 
\textit{"extraction of structured information from street view imagery"}. 
Hara et al.~\cite{hara2012feasibility,hara2013combining,hara2014tohme,hara2015improving} 
used \textit{"google street view using crowdsourcing, computer vision, and machine learning"} 
for an extensive set of detections and challenges related to street-level 
accessibility for persons with special needs. 
Frome et al.~\cite{frome2009large} presented methods for large-scale privacy 
protection in Google Street View imagery. Using their fully automatic system 
the authors were \textit{"able to sufficiently blur more than 89\% of faces and 94--96\% of 
license plates in evaluation sets sampled from Google Street View imagery"}.
Related to the accurate mapping of an image based on street-level imagery, 
Zamir and Shah~\cite{zamir2010accurate} introduced methods for image 
localization using Google Street View with an accuracy comparable to 
GPS-based technology. 

However, none of the existing works attempted to perform detection and 
accurate location mapping of CCTV camera objects from street-level imagery. 
In this context, our work is the first to achieve this and perfectly extends 
the state of the art by accurately detecting CCTV cameras on both street-level 
and other imagery (both geotagged and not). 

\subsection{Face recognition, privacy and CCTV}
\label{sec:facerec}
Face recognition is a very hot topic, has been a prolific area of research, 
and there is an immense body of works~\cite{turk1991face,bruce1986understanding,brunelli1993face,ahonen2004face,jain2011handbook,phillips2005overview,parkhi2015deep}. 
At the same time, (near) real-time face detection, tracking, and recognition 
is one of the core applications of CCTV cameras which is increasingly gaining 
traction. 
Several papers have been released on the subject of facial detection and 
recognition in real-time recordings such as the ones in CCTV setups. 
%
Halawa et al. \cite{halawa_et_al} recently showed how the 
\textit{Faster R-CNN}~\cite{ren_faster} algorithm is used with face 
detection in CCTV systems. 
%
Bah and Ming~\cite{bah/ming} presented their facial recognition algorithm. 
Their method includes prepossessing the images to better capture facial 
features and then implementing their own Local Binary Pattern (LBP) 
algorithm~\cite{ahonen2004face}. 
%
Mileva and Burton~\cite{milewa_burton} demonstrated the use of facial 
recognition in a noisy real-life environment such as in CCTV camera footage. 
The authors conclude that facial recognition from CCTV footage is certainly 
an arduous task and prone to errors~\cite{milewa_burton}. 
Worse, erroneous face recognition results may lead to arrests of innocent 
persons, multiple times the arrests were found to be wrong~\cite{losingface16}. 

However, large CCTV networks may provide unaccountable access to their 
video streams, turning them into tools of \emph{illegal mass surveillance}~\cite{moscow2019cctv}. 
Additionally, the researchers were able for a fee to get access 
to the \emph{``face search''} feature, where the requestor provides a photo 
containing a ``reference face''. Then, the facial recognition 
sub-system of the CCTV network provides back an extensive list of 
exact geo-locations (and other meta information) where similarly 
matching faces were previously seen within the CCTV network. 
The access to video feeds and to the \emph{``face search''} feature is allegedly 
sold for a very low fee on underground and specialized forums by 
unscrupulous law enforcement officers (i.e., a particular instance 
of the \textit{insider threat}~\cite{bishop2008defining,spitzner2003honeypots}). 

In this context, our research, through the automated CCTV camera detection 
(having as immediate goal accurate mapping of camera geo-locations and characteristics) 
aims to provide users with tools for a more democratic use of technology 
where privacy controls are on the users' side. For example, such tools may 
help the users to make an informed and real-time decision whether they want 
to be (or not!) in an area within the field of view of CCTV cameras.

\subsection{Cybersecurity and CCTV/IP cameras}

Recent research demonstrated that the state of cybersecurity for IoT 
devices (including DVRs and CCTV/IP cameras) and their firmware is very 
bad~\cite{cui2010quantitative,costin2014large,costin2016automated}.
Back in 2013, independently and almost simultaneously 
Heffner~\cite{heffner2013exploiting}, 
Costin~\cite{costin2013poor}, 
Shekyan and Harutyunyan~\cite{shekyan2013watch} 
researched and presented about vulnerabilities, exploits and cybersecurity 
dangers related to vulnerable and exposed CCTV/IP/surveillance cameras. 
In particular,~\cite{costin2013poor} proposed several 
hypotheses related to (in)security of CCTV/IP cameras, which were later 
extended and systematized in~\cite{costin2016security}.
%
%
Unsurprisingly, in 2013--2014 reports started to surface about the 
infamous (and supposedly Russian-operated) Insecam 
project~\cite{insecam2013,xu2018internet}, which at its peak featured 
between 100-200k video feeds from vulnerable/insecure CCTV/IP cameras 
connected to the internet.
%
Moreover, in late 2016 news broke about the infamous and devastating 
Mirai IoT botnet~\cite{krebs2016source,antonakakis2017understanding}.
It featured the largest known DDoS attacks to date of over 1 TB/s, 
employed at peak about 600k devices, took down major parts of the internet, 
and mostly consisted of hacked CCTV/IP cameras and CCTV/DVR surveillance systems. 

While this work does not explicitly address the cybersecurity of CCTV cameras,
a possible offensive scenario is presented in Section~\ref{sec:ethic}. 
%



\section{Conclusion}
\label{sec:concl}

In this paper, we presented the first computer vision object 
detectors aiming to accurately identify CCTV cameras in images and video frames. 
%
%
To build our system, we used and evaluated in parallel several 
state-of-the-art computer vision frameworks and backbones. 
%
%
To this end, our best detectors were built using \DSAllTrainCountAll{} 
images that were manually reviewed and annotated to contain 
\DSAllTrainCountInstancesAll{} CCTV camera instances, 
and achieve an accuracy of up to \DetValEightAP{}. 

Furthermore, we develop and introduce by the way of preview and main excerpts, 
the first known and working prototypes of:
%
a) \emph{CCTV-aware route planning and navigation};
b) \emph{CCTV-aware mobile devices} (data collection and real-time mapping);
c) \emph{Browser-only image annotation toolset}.
%
These prototypes are motivated and powered by the core object detectors effort 
presented above.
%
We presented excerpts and proofs of working prototypes, however we leave 
their detailed presentation as separate publications. 

Finally, with this work, we hope to motivate in several ways the 
communities of researchers, practitioners, policy-makers, and end-users. 
First and more general is to focus the debates and policy-making related 
to privacy and safety of CCTV cameras towards a more science-, technology- 
and research-driven ground. 
Second is to encourage the improvement of our object detectors and techniques 
presented, so that they can be immediately incorporated at a larger scale, 
both in cloud and edge applications, and for the greater privacy benefit. 
With these in mind, we release the relevant artefacts (e.g., code, datasets, trained models, and documentation) at: 
\textit{\url{https://github.com/Fuziih}} and 
\textit{\url{https://etsin.fairdata.fi/dataset/d2d2d6e2-0b5c-46e0-8833-53d8a24838a0} (urn:nbn:fi:att:258ce5ad-9501-46b9-a707-c1f59689ee10)}.


\section*{Acknowledgments}

We acknowledge grants of computer capacity from the 
Finnish Grid and Cloud Infrastructure (FGCI) 
(persistent identifier \texttt{urn:nbn:fi:research-infras-2016072533}).
Part of this research was kindly supported by the 
\emph{``17.06.2020 Decision of the Research Dean on research funding within the faculty''} 
grant from the Faculty of Information Technology of the University of Jyv\"askyl\"a, 
and the grant was facilitated and managed by Dr. Andrei Costin.
Authors would also like to thank the following persons for their dedicated 
efforts during the crowdsource data collection phase: 
Pyry Kotilainen,
Janne Uusitupa, 
Arttu Takala, 
Syed Khandker, 
Anna Arikainen. 

Hannu Turtiainen would like to also thank the Finnish Cultural Foundation / Suomen Kulttuurirahasto (https://skr.fi/en) 
for supporting his PhD dissertation work and research (grant decision 00211119), and the Faculty of Information Technology of University of Jyv\"{a}skyl\"{a} (JYU), 
in particular Prof. Timo H\"{a}m\"{a}l\"{a}inen, for partly supporting his PhD supervision at JYU during 2021--2022.



\clearpage


\bibliographystyle{IEEEtran}
\bibliography{paper}


\appendix


\section{Appendix: Additional Figures}

\begin{figure}[htb]
    \centering
    \includegraphics[width=0.8\columnwidth]{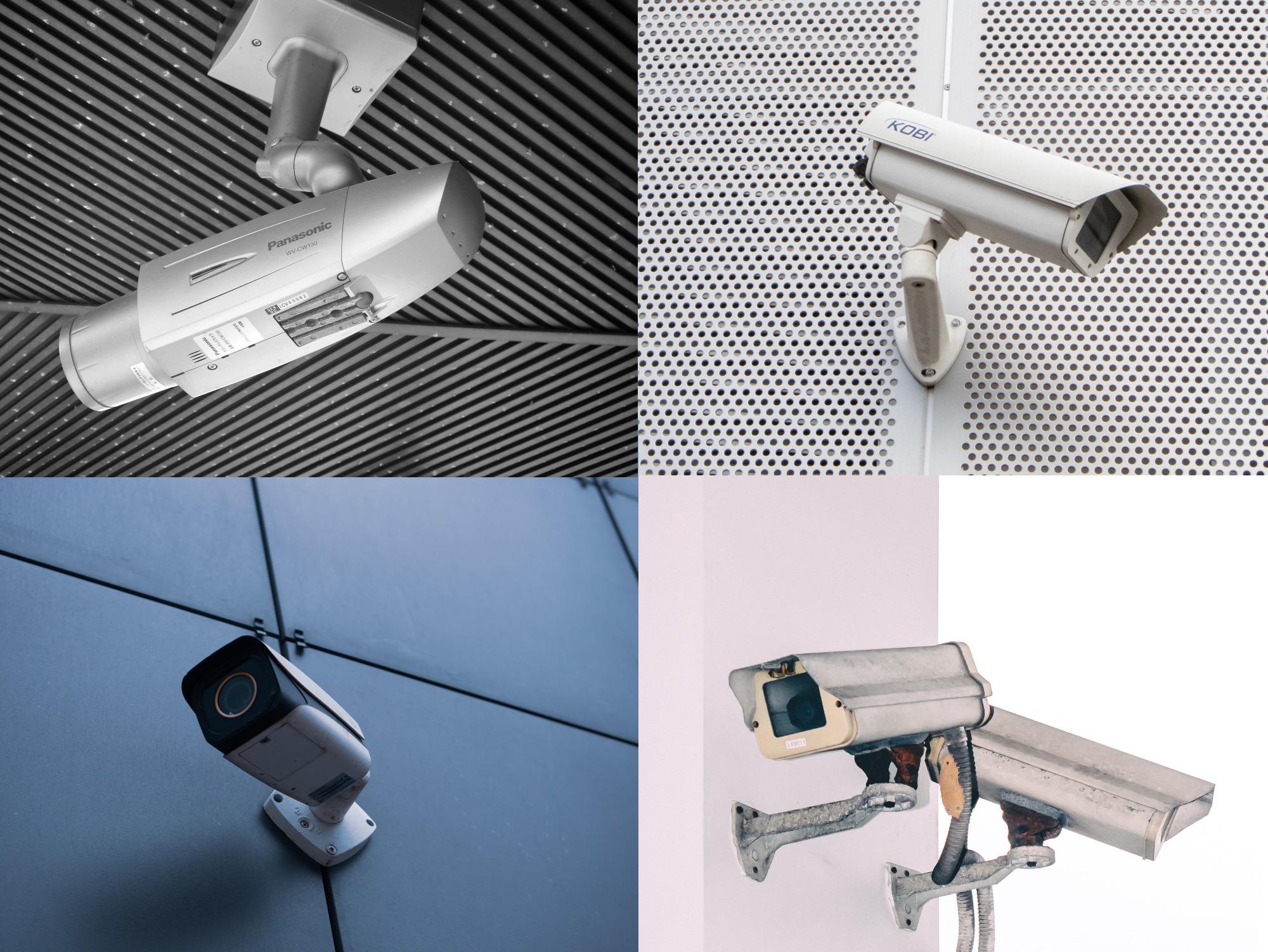}
    \caption{Examples of \emph{directed camera} class (images CC0 by \url{unsplash.com}).} 
    \label{fig:sample_directed}
\end{figure}

\begin{figure}[htb]
    \centering
    \includegraphics[width=0.8\columnwidth]{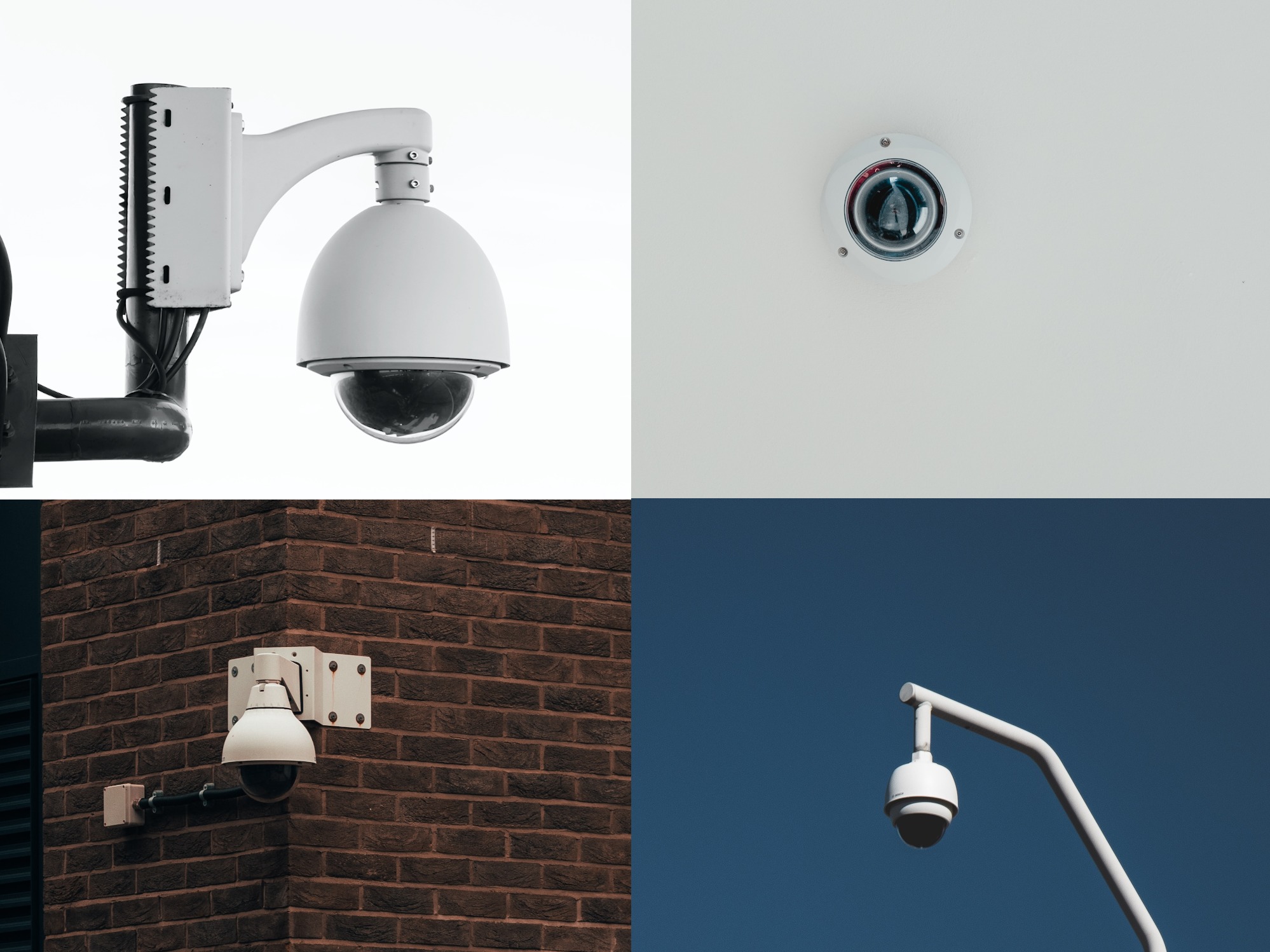}
    \caption{Examples of \emph{round camera} class (images CC0 by \url{unsplash.com}).} 
    \label{fig:sample_round}
\end{figure}

\begin{figure}[htb]
  \centering
  \includegraphics[width=0.8\columnwidth]{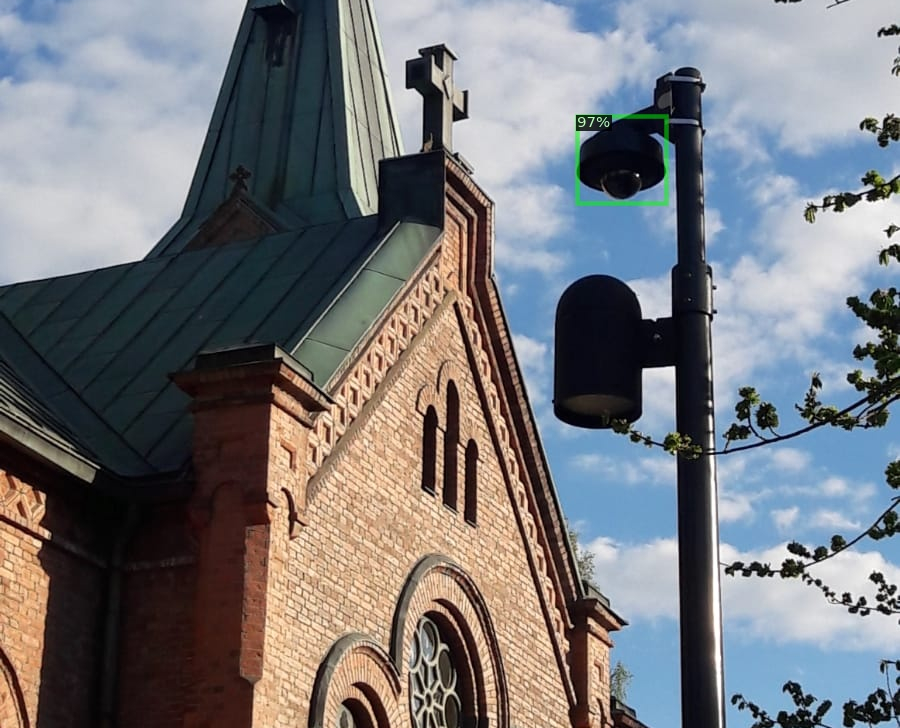}
  \caption{Example of a hybrid installation consisting of a light pole combined with a CCTV camera. Our TridentNet model works well with a detection confidence of 97\% of the ``round class'' instance, and avoids the potential ``directed class'' FP represented by the light fixture below the camera.}
  \label{fig:20200527}
\end{figure}

\begin{figure}[htb]
    \centering
    \includegraphics[width=0.8\columnwidth]{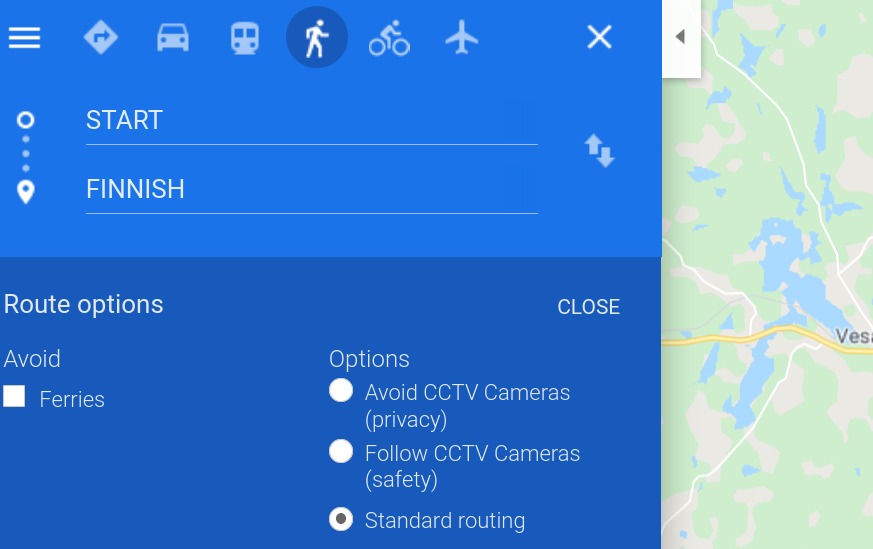}
    \caption{Our proposed vision for a novel map navigation that provides the 
    users both privacy (\emph{``Avoid CCTV Cameras (privacy-first)''}) 
    and safety (\emph{``Follow CCTV Cameras (safety-first)''}) route planning options.}
    \label{fig:routes-gmaps}
\end{figure}

\begin{figure}[htb]
    \centering
    \includegraphics[width=0.8\columnwidth]{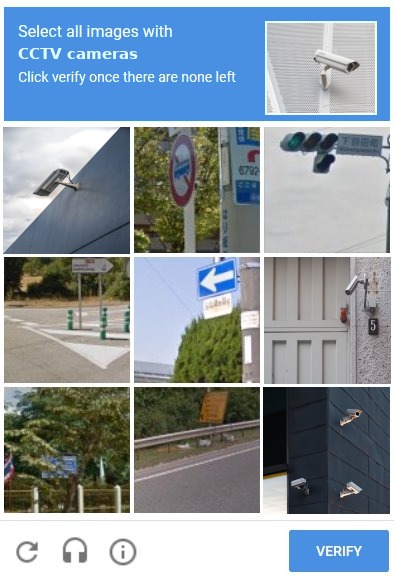}
    \caption{Our proposed vision for a novel reCAPTCHA V2 extension offering \emph{``Select all images with \textbf{CCTV cameras}''} to better and faster help annotate and validate the CCTV camera objects in Google Street View imagery.}
    \label{fig:idea-recaptchav2}
\end{figure}

\begin{figure}[htb]
    \centering
    \includegraphics[width=0.8\columnwidth]{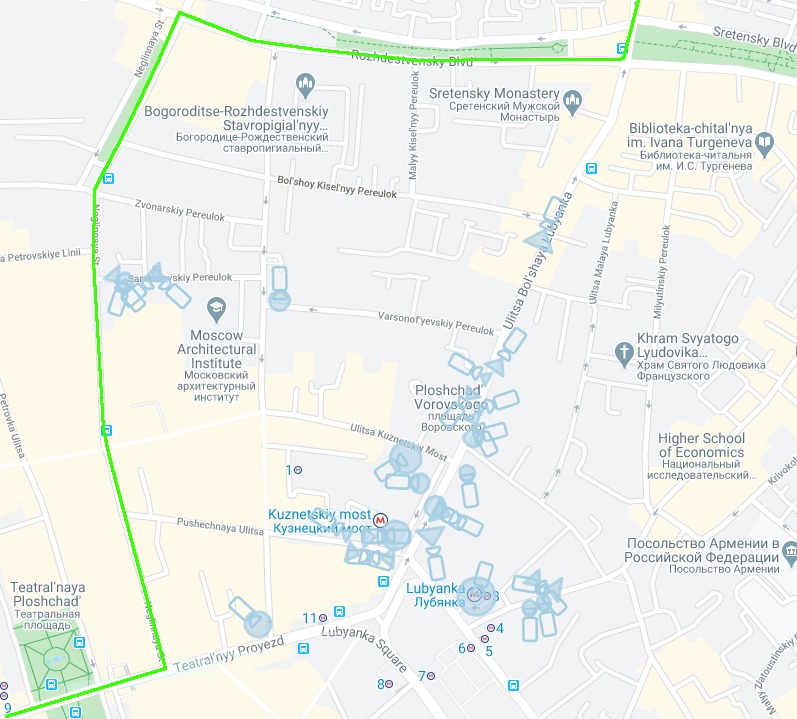}
    \caption{Google Maps prototype: CCTV-aware routing system providing route for 
    \emph{``Avoid CCTV Cameras (privacy-first)''} user option.
    }
    \label{fig:routes-good-gmaps}
\end{figure}

\begin{figure}[htb]
    \centering
    \includegraphics[width=0.8\columnwidth]{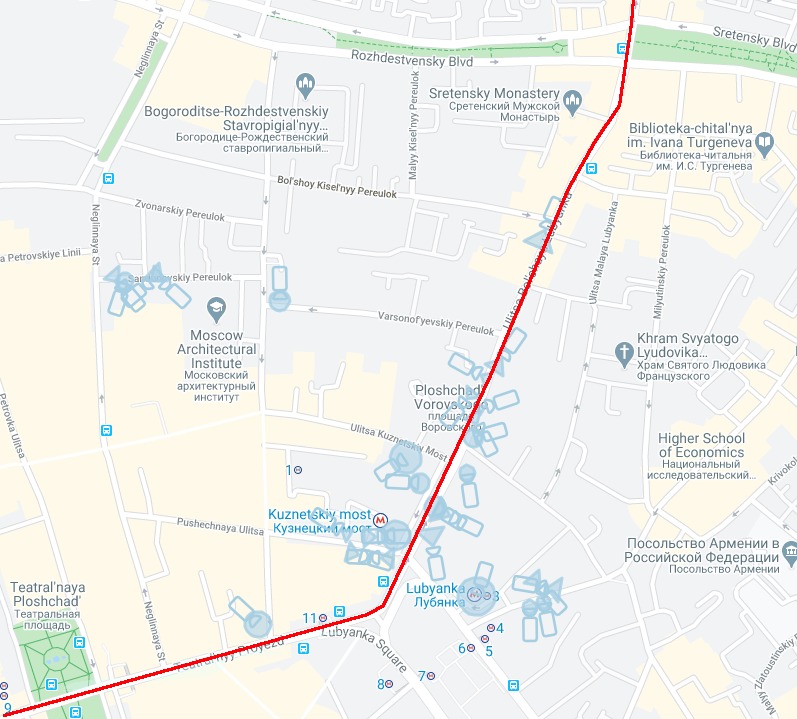}
    \caption{Google Maps prototype: CCTV-aware routing system providing route for 
    \emph{``Follow CCTV Cameras (safety-first)''} user option.
    }
    \label{fig:routes-bad-gmaps}
\end{figure}

\begin{figure}
    \centering
    \includegraphics[width=1.0\columnwidth]{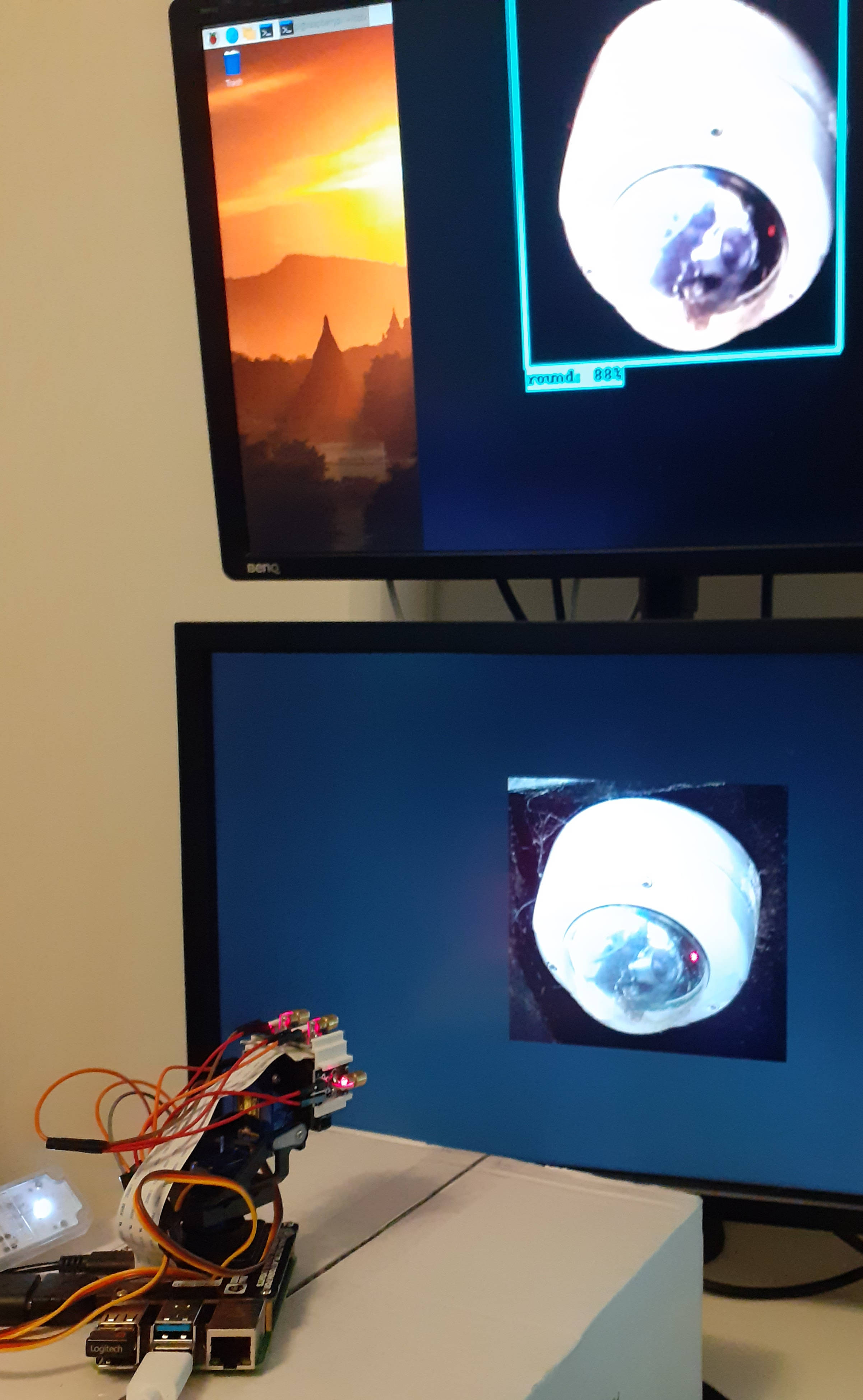}
    \caption{
    Mobile device prototype in action -- camera detected, lasers activated. Top screen shows the RPi4 internal view. Bottom screen shows a CCTV camera instance to test the mobile device.
    }
    \label{fig:rpi4_pantilt_coral}
\end{figure}

\begin{algorithm}[htb]
\caption{CCTV-aware mobile device operation}
\label{lst:mobiledev}
\begin{algorithmic}[1]
\renewcommand{\algorithmicrequire}{\textbf{Input:}}
\renewcommand{\algorithmicensure}{\textbf{Output:}}
\REQUIRE --
\ENSURE location and meta data of CCTV cameras
\\ 
\textit{INIT} :
\STATE START autonomous operation
\\ 
\WHILE {(TRUE)}
\STATE SCAN with Pan-Tilt HAT with PiCamera street level imagery
\IF {(CCTV camera object IS detected)}
\STATE CENTER with Pan-Tilt HAT both range finder and PiCamera on the detected CCTV camera
\STATE \textbf{OPTIONAL}: TRIGGER special lasers for camera-blinding ("pro-active privacy" approach)
\STATE TRIGGER PiCamera
\STATE CAPTURE with PiCamera visual sample for data validation (e.g., reCAPTCHA in Section~\ref{sec:annot})
\STATE TRIGGER laser-based range finder
\STATE CAPTURE with range finder accurate distance from the mobile device to the detected CCTV camera
\STATE COMBINE range finder distance with accurate GPS location from the GPS dongle
\STATE UPDATE/ADD CCTV camera meta-data (visual, exact location) to the global real-time map
\STATE SAVE CCTV camera data / SEND to API cloud
\ENDIF
\ENDWHILE
\RETURN $CCTV\_DATA$ 
\end{algorithmic} 
\end{algorithm}


\onecolumn

\section{Appendix: Our system applied to a real-life experiment on CCTV and privacy by Pasley~\cite{bi2019cctv}}
\label{sec:apdx1}

\begin{figure*}[htb]
    \centering
    \includegraphics[width=1.00\textwidth]{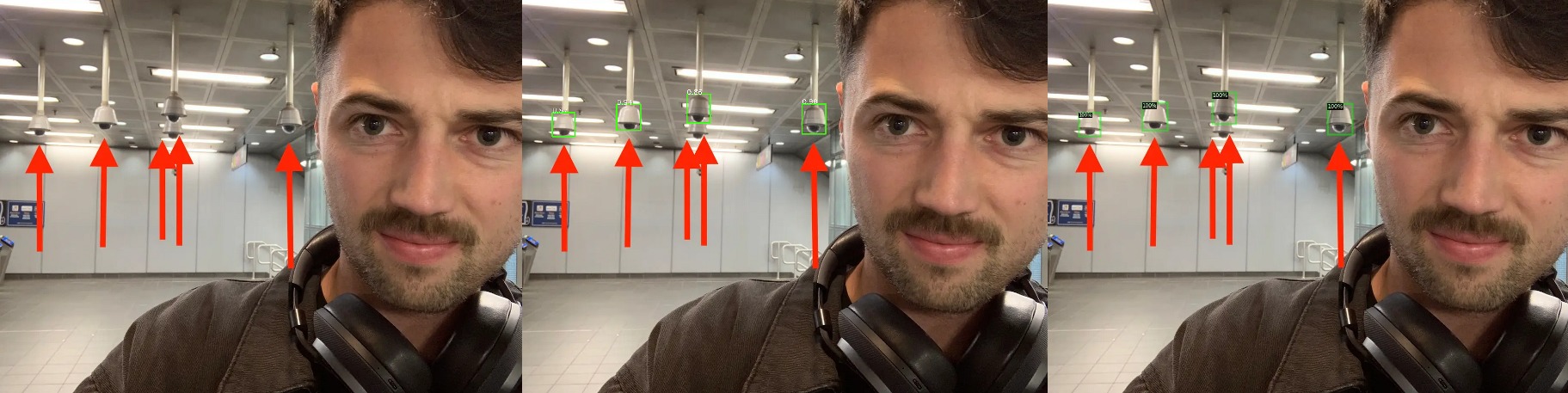}
    \caption{Visual results (left to right): Original image (Ground Truth) - 5 TP; ATSS X-101 - 4 TP, 1 FN; TridentNet R-101 - 4 TP, 1 FN}
    \label{fig:bi_1v2}
\end{figure*}

\begin{figure*}[htb]
    \centering
    \includegraphics[width=1.00\textwidth]{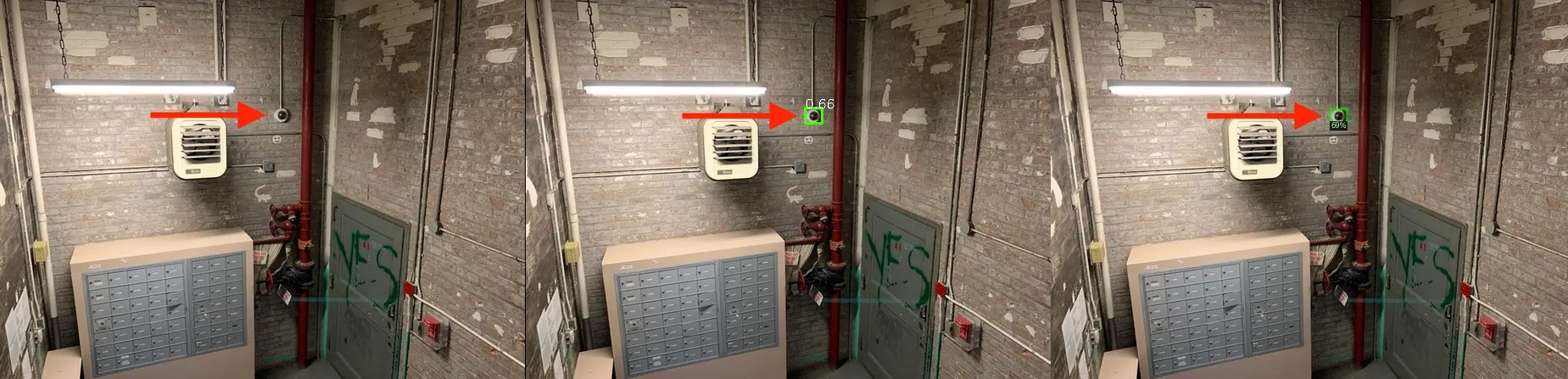}
    \caption{Visual results (left to right): Original image (Ground Truth) - 1 TP; ATSS X-101 - 1 TP; TridentNet R-101 - 1 TP}
    \label{fig:bi_5}
\end{figure*}

\begin{figure*}[htb]
    \centering
    \includegraphics[width=1.00\textwidth]{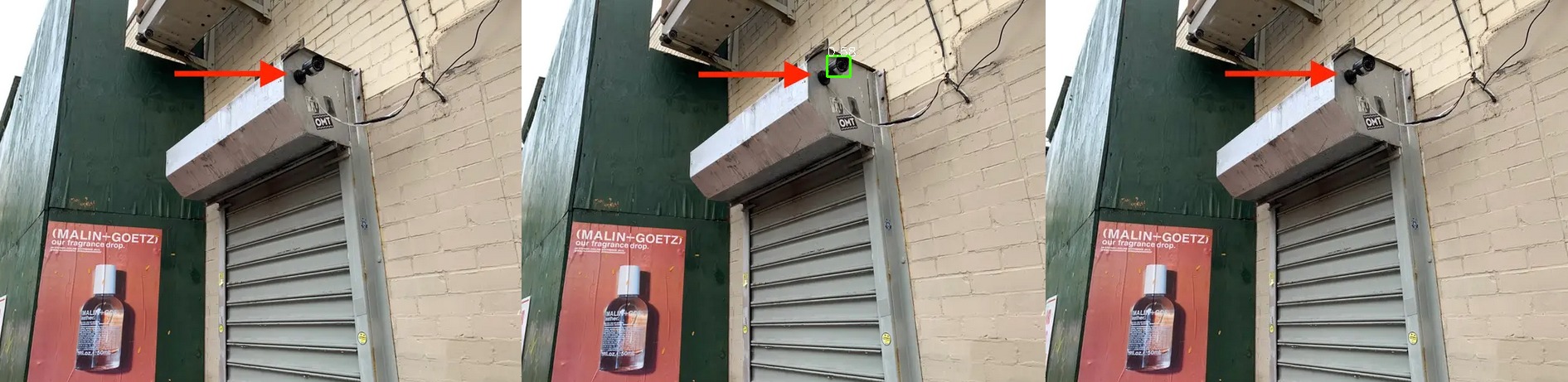}
    \caption{Visual results (left to right): Original image (Ground Truth) - 1 TP; ATSS X-101 - 1 TP; TridentNet R-101 - 1 FN}
    \label{fig:bi_11}
\end{figure*}

\begin{figure*}[htb]
    \centering
    \includegraphics[width=1.00\textwidth]{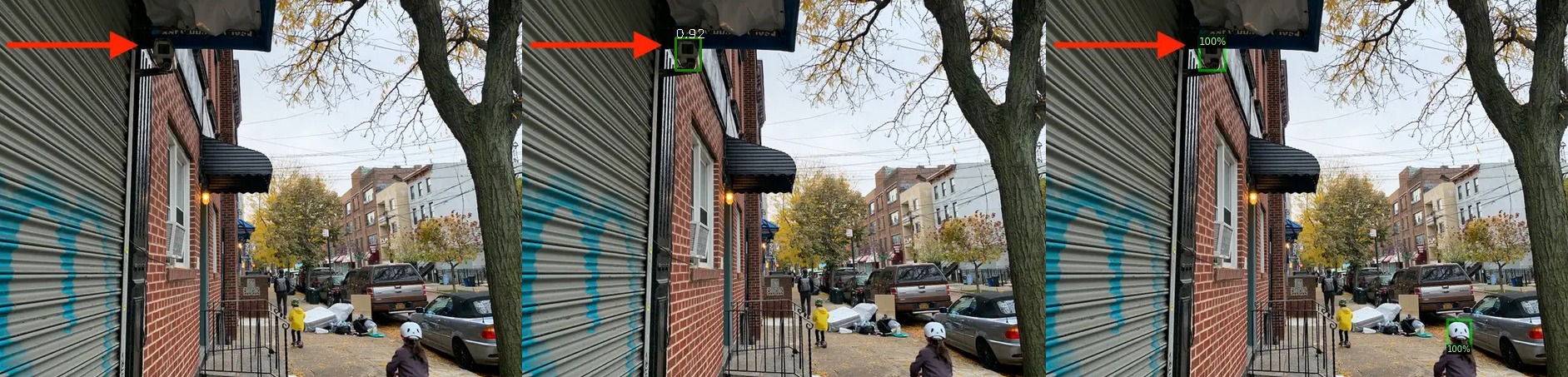}
    \caption{Visual results (left to right): Original image (Ground Truth) - 1 TP; ATSS X-101 - 1 TP; TridentNet R-101 - 1 TP, 1 FP}
    \label{fig:bi_13}
\end{figure*}

\begin{figure*}[htb]
    \centering
    \includegraphics[width=1.00\textwidth]{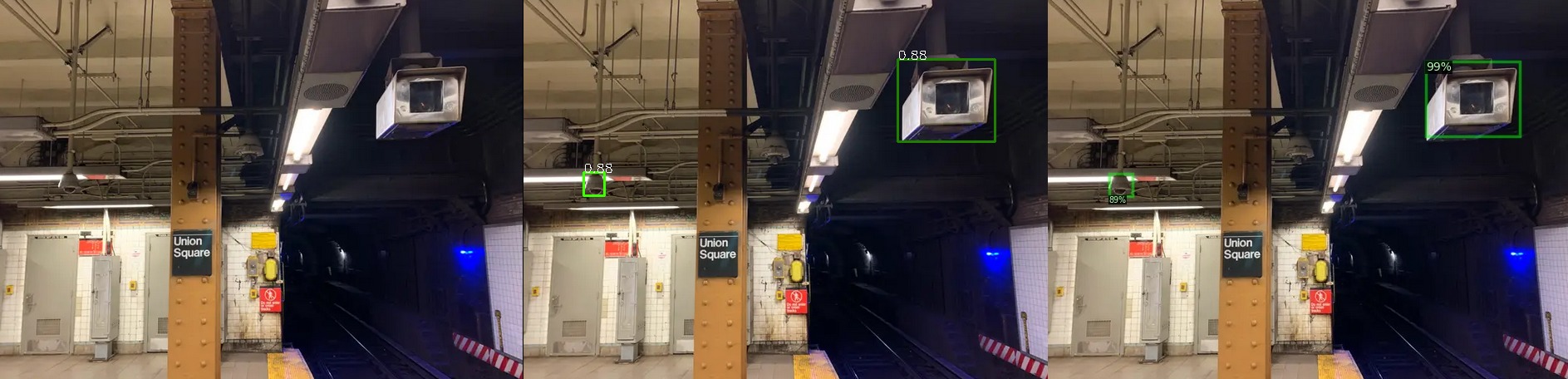}
    \caption{Visual results (left to right): Original image (Ground Truth) - 2 TP; ATSS X-101 - 2 TP; TridentNet R-101 - 2 TP}
    \label{fig:bi_16}
\end{figure*}

\begin{figure*}[htb]
    \centering
    \includegraphics[width=1.00\textwidth]{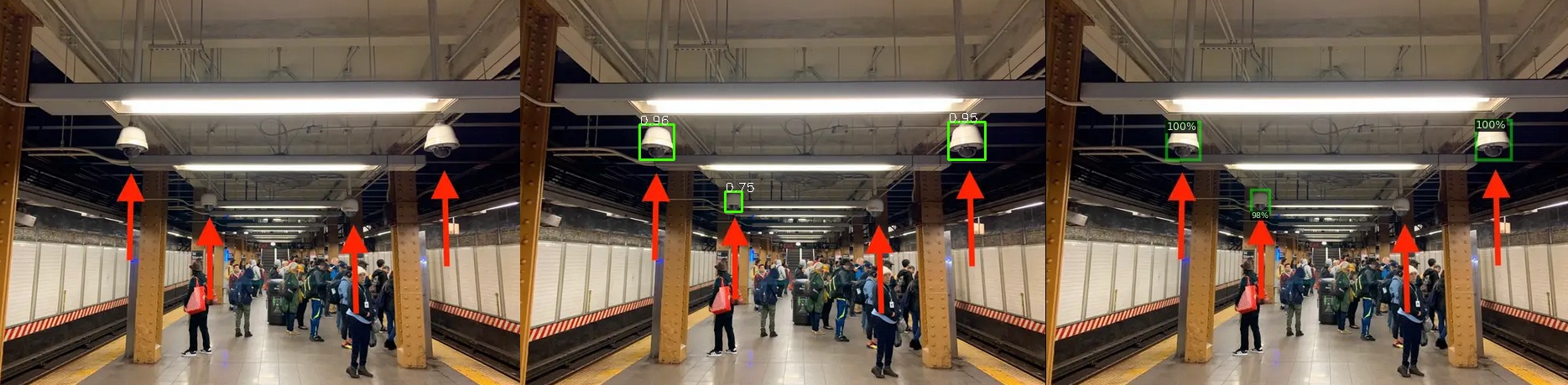}
    \caption{Visual results (left to right): Original image (Ground Truth) - 4 TP; ATSS X-101 - 3 TP, 1 FN; TridentNet R-101 - 3 TP, 1 FN}
    \label{fig:bi_17}
\end{figure*}

\begin{figure*}[htb]
    \centering
    \includegraphics[width=1.00\textwidth]{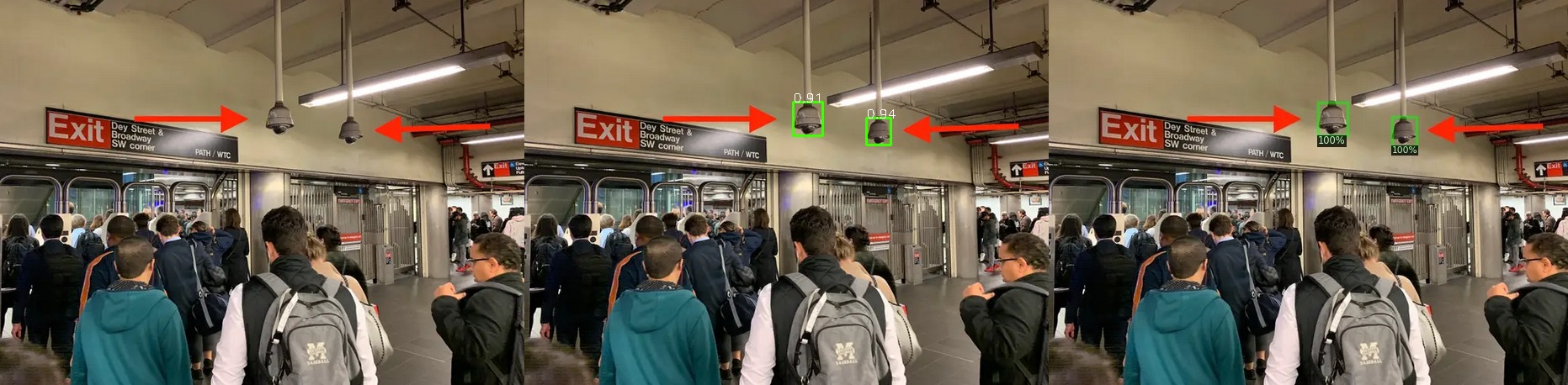}
    \caption{Visual results (left to right): Original image (Ground Truth) - 2 TP; ATSS X-101 - 2 TP; TridentNet R-101 - 2 TP}
    \label{fig:bi_19}
\end{figure*}

\begin{figure*}[htb]
    \centering
    \includegraphics[width=1.00\textwidth]{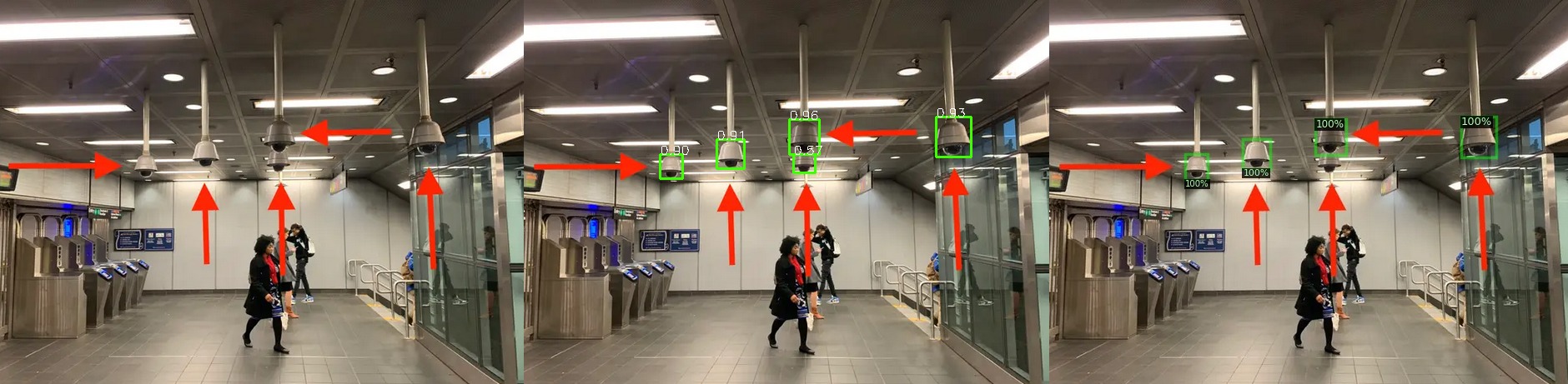}
    \caption{Visual results (left to right): Original image (Ground Truth) - 5 TP; ATSS X-101 - 5 TP; TridentNet R-101 - 4 TP, 1 FN}
    \label{fig:bi_20}
\end{figure*}

\begin{figure*}[htb]
    \centering
    \includegraphics[width=1.00\textwidth]{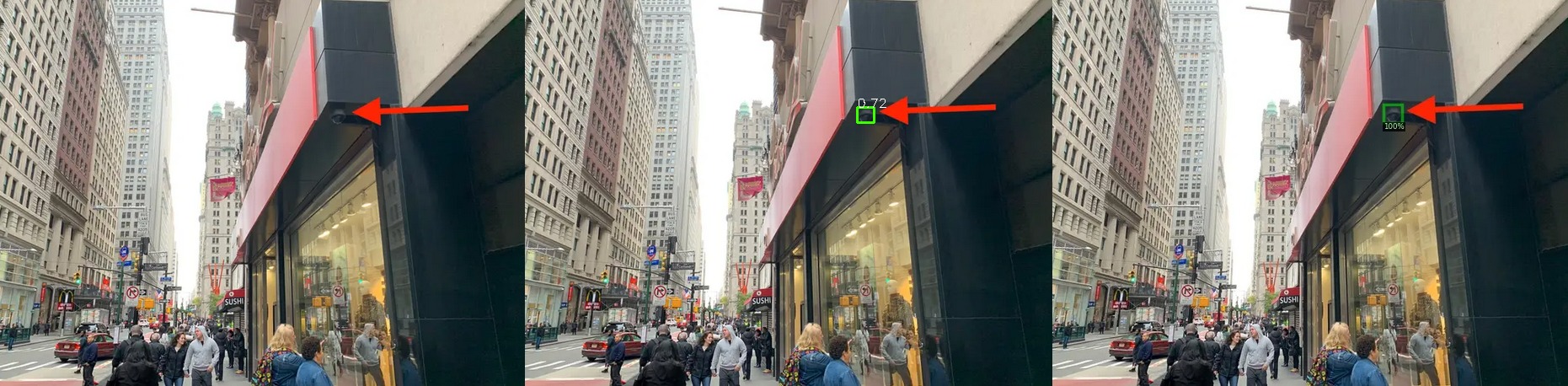}
    \caption{Visual results (left to right): Original image (Ground Truth) - 1 TP; ATSS X-101 - 1 TP; TridentNet R-101 - 1 TP}
    \label{fig:bi_21}
\end{figure*}

\begin{figure*}[htb]
    \centering
    \includegraphics[width=1.00\textwidth]{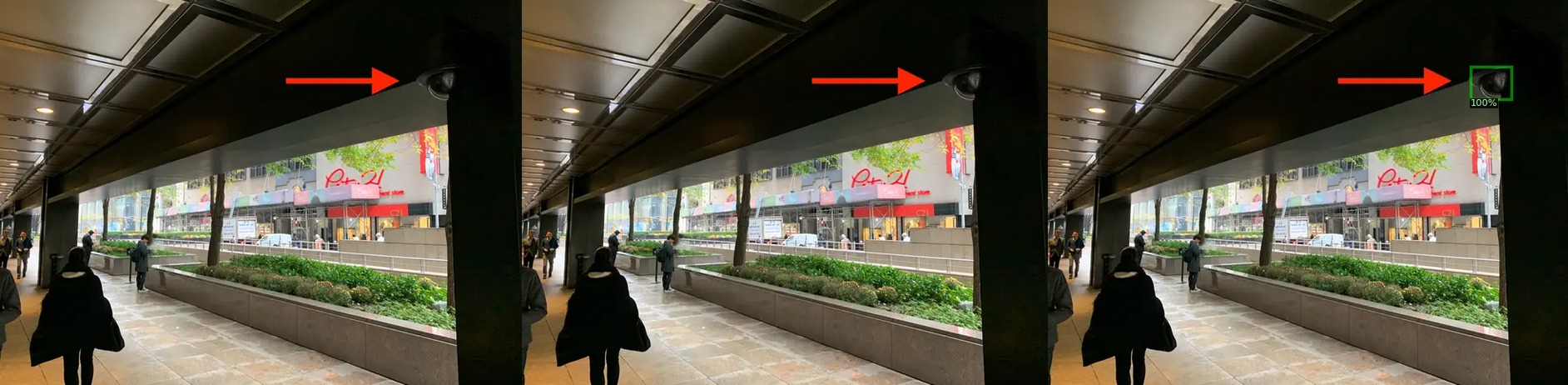}
    \caption{Visual results (left to right): Original image (Ground Truth) - 1 TP; ATSS X-101 - 1 FN; TridentNet R-101 - 1 TP. NOTE: When the original author's red arrow is removed, it proved quite challenging for humans to quickly detect the CCTV camera as it ``hides'' behind a corner and blends into the dark background.}
    \label{fig:bi_22}
\end{figure*}


\clearpage
\section{Appendix: Detailed Tables for \texttt{Dataset0}}
\label{sec:apdx-ds0}

\begin{table*}[htb]
\centering
\caption {Results for bounding box detection with the \texttt{Dataset0} \emph{testing set} (\textbf{bold}=best, \underline{underline}=worst).}
\label{tab:ds0-tab2}
  \begin{tabular}{ccccccccc}
    \toprule
    Detector & AP@0.5 & AP@0.5:0.95 & APs & APm & AR 1 & AR 10 & ARs & ARm \\
    \midrule
	CM2 Lite V-39 & \CmLiteAP & \CmLiteAPxx & \textbf{\CmLiteAPs} & \CmLiteAPm & \CmLiteARone & \textbf{\CmLiteARten} & \textbf{\CmLiteARs} & \CmLiteARm \\
	CM2 V-57 & \CmFiveSevenAP & \CmFiveSevenAPxx & \CmFiveSevenAPs & \CmFiveSevenAPm & \CmFiveSevenARone & \CmFiveSevenARten & \CmFiveSevenARs & \CmFiveSevenARm\\
	CM2 V-99 & \CmNineNineAP & \CmNineNineAPxx & \CmNineNineAPs & \textbf{\CmNineNineAPm} & \CmNineNineARone & \CmNineNineARten & \CmNineNineARs & \textbf{\CmNineNineARm} \\
	ATSS R-50 & \underline{\ATSSfiftyAP} & \underline{\ATSSfiftyAPxx} & \ATSSfiftyAPs &\underline{\ATSSfiftyAPm} & \ATSSfiftyARone & \ATSSfiftyARten & \ATSSfiftyARs & \underline{\ATSSfiftyARm} \\
	ATSS X-101 & \textbf{\ATSSxAP} & \textbf{\ATSSxAPxx} & \ATSSxAPs & \ATSSxAPm & \textbf{\ATSSxARone} & \ATSSxARten & \ATSSxARs & \ATSSxARm \\
	Trident R-101 & \TridentAP & \TridentAPxx & \underline{\TridentAPs} & \TridentAPm & \underline{\TridentARone} & \underline{\TridentARten} & \underline{\TridentARs} & \TridentARm \\
	\bottomrule
\end{tabular}
\end{table*}

\begin{table*}[htb]
\centering
\caption {Results for bounding box detection with the \texttt{Dataset0} \emph{validation set} (\textbf{bold}=best, \underline{underline}=worst).}
\label{tab:ds0-tab3}
  \begin{tabular}{ccccccccccc}
    \toprule
    Detector & AP@0.5 & AP@0.5:0.95 & APs & APm & APl & AR 1 & AR 10 & ARs & ARm & ARl \\
    \midrule
	CM2 Lite V-39 & \CmLiteVAP & \underline{\CmLiteVAPxx} & \underline{\CmLiteVAPs} & \CmLiteVAPm & \underline{\CmLiteVAPl} &\underline{\CmLiteVARone} & \underline{\CmLiteVARten} & \underline{\CmLiteVARs} & \underline{\CmLiteVARm} & \underline{\CmLiteVARl} \\
	CM2 V-57 & \textbf{\CmFiveSevenVAP} & \textbf{\CmFiveSevenVAPxx} & \textbf{\CmFiveSevenVAPs} & \CmFiveSevenVAPm & \CmFiveSevenVAPl & \textbf{\CmFiveSevenVARone} & \textbf{\CmFiveSevenVARten} & \textbf{\CmFiveSevenVARs} & \CmFiveSevenVARm & \CmFiveSevenVARl \\
	CM2 V-99 & \CmNineNineVAP & \CmNineNineVAPxx & \CmNineNineVAPs & \CmNineNineVAPm & \textbf{\CmNineNineVAPl} & \CmNineNineVARone & \CmNineNineVARten & \CmNineNineVARs & \CmNineNineVARm & \textbf{\CmNineNineVARl} \\
	ATSS R-50 & \underline{\ATSSfiftyVAP} & \ATSSfiftyVAPxx & \ATSSfiftyVAPs & \ATSSfiftyVAPm & \ATSSfiftyVAPl & \ATSSfiftyVARone & \ATSSfiftyVARten & \ATSSfiftyVARs & \ATSSfiftyVARm & \ATSSfiftyVARl \\
	ATSS X-101 & \ATSSxVAP & \ATSSxVAPxx & \ATSSxVAPs & \textbf{\ATSSxVAPm} & \ATSSxVAPl &\ATSSxVARone & \ATSSxVARten & \ATSSxVARs & 	\textbf{\ATSSxVARm} & \ATSSxVARl \\
Trident R-101 & \TridentVAP & \TridentVAPxx & \TridentVAPs & \underline{\TridentVAPm} & \TridentVAPl &\TridentVARone & \TridentVARten & \TridentVARs & \TridentVARm & \TridentVARl \\
	\bottomrule
\end{tabular}
\end{table*}

\begin{table*}
\centering
\caption {Detector configuration, iterations count, training and inference times when training detectors on \texttt{Dataset0}. } 
\label{tab:ds0-tab1}
  \begin{tabular}{lcccccc}
    \toprule
    Detector & \specialcell{Best-result \\ iterations \\ (number)} & \specialcell{Weights \\ file size \\ (MB)} & \specialcell{Avg. train \\ time / iter. \\ (seconds)} & \specialcell{Avg. inference \\ time / image \\ (seconds)} \\
    \midrule
    \specialcell{CM2 Lite V-39} & \CmLiteIter & \CmLiteFile & \textbf{\CmLiteTrainTime} & \textbf{\CmLiteInf} \\
	\specialcell{CM2 V-57} & \CmFiveSevenIter & \CmFiveSevenFile & \CmFiveSevenTrainTime & \CmFiveSevenInf \\
	\specialcell{CM2 V-99} & \CmNineNineIter & \underline{\CmNineNineFile} & \CmNineNineTrainTime & \CmNineNineInf \\
	\specialcell{ATSS R-50} & \ATSSfiftyIter & \textbf{\ATSSfiftyFile} & \ATSSfiftyTrainTime & \ATSSfiftyInf \\
	\specialcell{ATSS X-101} & \ATSSxIter & \ATSSxFile & \underline{\ATSSxTrainTime} & \ATSSxInf \\
	\specialcell{Trident R-101} & \TridentIter & \TridentFile & \TridentTrainTime & \underline{\TridentInf} \\
	\bottomrule
  \end{tabular}%
\end{table*}



\end{document}